\documentclass[lettersize,journal]{IEEEtran}
\usepackage{amsmath,amsfonts}
\usepackage{algorithmic}
\usepackage{algorithm}
\usepackage{array}
\usepackage{textcomp}
\usepackage{stfloats}
\usepackage{url}
\usepackage{verbatim}
\usepackage{graphicx}
\usepackage{diagbox}
\usepackage{cite}
\usepackage{amsthm}
\usepackage{picinpar}
\usepackage{amssymb}
\usepackage{colortbl}
\usepackage{soul}
\usepackage{multirow}
\usepackage{pifont}
\usepackage{bbding}
\usepackage{color}
\usepackage{alltt}
\usepackage{enumerate}
\usepackage{siunitx}
\usepackage{breakurl}
\usepackage{epstopdf}
\usepackage{abraces}
\usepackage{pbox}
\usepackage{bm} 
\usepackage{booktabs} 
\usepackage{threeparttable} 
\usepackage{float} 
\usepackage{empheq}
\usepackage{subfigure}
\usepackage{cases}
\usepackage{hyperref}
\newtheorem{assumption}{Assumption}
\newtheorem{problem}{Problem}
\newtheorem{remark}{Remark}
\newtheorem{definition}{Definition}

\newtheorem{theorem}{Theorem}

\begin{document}

\title{InPTC: Integrated Planning and Tube-Following Control for Prescribed-Time Collision-Free Navigation of Wheeled Mobile Robots}

\author{Xiaodong Shao,~\IEEEmembership{Member,~IEEE,} 
Bin Zhang, Hui Zhi,
Jose Guadalupe Romero,~\IEEEmembership{Member,~IEEE,}
Bowen Fan,
Qinglei Hu,~\IEEEmembership{Senior Member,~IEEE,} and 
David Navarro-Alarcon,~\IEEEmembership{Senior Member,~IEEE}
\thanks{This work was supported in part by the Research Grants Council (RGC) of Hong Kong under grants 15212721 and 15231023. \textit{(X. Shao and B. Zhang contributed equally.) (Corresponding author: David Navarro-Alarcon)}.}        
\thanks{X. Shao and Q. Hu are with the School of Automation Science and Electrical Engineering, Beihang University, Beijing 100191, China (e-mail: xdshao\_sasee@buaa.edu.cn; huql\_buaa@buaa.edu.cn). B. Zhang, H. Zhi, B. Fan, and D. Navarro-Alarcon are with the Department of Mechanical Engineering, The Hong Kong Polytechnic University (PolyU), Kowloon, Hong Kong (e-mail: me-bin.zhang@connect.polyu.hk, hui1225.zhi@connect.polyu.hk, bo-wen.fan@connect.polyu.hk, dnavar@polyu.edu.hk). J. G. Romero is with the Departamento Académico de Ingeniería Eléctrica y Electrónica, Instituto Tecnológico Autónomo de México (ITAM), Mexico City, Mexico (email: jose.romerovelazquez@itam.mx).}
}

\markboth{Journal of \LaTeX\ Class Files,~Vol.~14, No.~8, August~2021}%
{Shell \MakeLowercase{\textit{et al.}}: A Sample Article Using IEEEtran.cls for IEEE Journals}


\maketitle

\begin{abstract}
In this article, we propose a novel approach, called InPTC (Integrated Planning and Tube-Following Control), for prescribed-time collision-free navigation of wheeled mobile robots in a compact convex workspace cluttered with static, sufficiently separated, and convex obstacles. A path planner with prescribed-time convergence is presented based upon Bouligand's tangent cones and time scale transformation (TST) techniques, yielding a continuous vector field that can guide the robot from almost all initial positions in the free space to the designated goal at a prescribed time, while avoiding entering the obstacle regions augmented with safety margin. By leveraging barrier functions and TST, we further derive a tube-following controller to achieve robot trajectory tracking within a prescribed time less than the planner's settling time. This controller ensures the robot moves inside a predefined ``safe tube'' around the reference trajectory, where the tube radius is set to be less than the safety margin. Consequently, the robot will reach the goal location within a prescribed time while avoiding collision with any obstacles along the way. The proposed InPTC is implemented on a Mona robot operating in an arena cluttered with obstacles of various shapes. Experimental results demonstrate that InPTC not only generates smooth collision-free reference trajectories that converge to the goal location at the preassigned time of $250\,\rm s$ (i.e., the required task completion time), but also achieves tube-following trajectory tracking with tracking accuracy higher than $0.01\rm m$ after the preassigned time of $150\,\rm s$. This enables the robot to accomplish the navigation task within the required time of $250\,\rm s$.
\end{abstract}


\section{Introduction}

\IEEEPARstart{T}{he} motion control of wheeled mobile robots (WMRs) is a benchmark problem in robotics due to its key role in extensive real-world applications, such as cargo transportation, automated patrolling, and exploration of hazardous environments \cite{klancar2017wheeled}. In these applications, WMRs typically operate in obstacle-cluttered environments, which motivates the development of safe navigation algorithms that can steer WMRs from an initial position to a desired goal location without colliding with any obstacles along the way \cite{chu2022feedback}. 
Collision-free navigation of WMRs have garnered significant attention from the robotics and control research communities \cite{lumelsky2005sensing}. 

Existing solutions to the robot navigation problem can be primarily classified into two categories \cite{berkane2021navigation}: global (map-based) methods and local (reactive) methods. The former requires a priori information of environments (e.g., position and shape of obstacles), while the latter utilizes only local knowledge of obstacles obtained from on-board sensors. 
Among the global methods, a computationally efficient solution is the artificial potential field (APF)-based approach \cite{khatib1986real}, which uses a field of potential forces to push the robot towards a goal position and pull it away from obstacles \cite{kim1992real,valbuena2012hybrid,li2021optimization}. Control barrier function (CBF) combined with quadratic programming (CBF-QP) also provides an effective strategy for developing controllers that can both stabilize a system and ensure safety, making it particularly useful for autonomous vehicles operating in obstacle-cluttered environments \cite{singletary2021comparative,srinivasan2020control,chen2020guaranteed}. Note that both the APF- and CBF-based methods often reach local minima, which hinders achieving global convergence to the designated goal, especially in topologically complex settings. Although navigation functions presented in \cite{rimon1992exact} can overcome this issue, they require unknown tuning of parameters. Other global methods include heuristic approaches such as $A^{\star}$ \cite{erke2020improved}, rapidly exploring random tree (RRT) \cite{noreen2016optimal}, and genetic algorithms \cite{tu2003genetic}. Since these methods are search-based solutions, their performance is directly affected by the problem scale \cite{costa2019survey}. Optimization-based methods \cite{yan2022hierarchical} provide alternatives, but typically require numerically solving constrained optimization problems, which results in high computational costs. Local methods that offer reactive solutions for collision-free robot motions are highly desirable, particularly in autonomous exploration applications, where robots often have limited access to environment information and must rely on local sensory data to detect obstacles. 
Bug algorithm is a simple reaction planning approaches for mobile robots \cite{lumelsky1990incorporating,mcguire2019comparative}. Arslan and Koditschek \cite{arslan2019sensor} proposed a reactive method based on a robot-centric spatial decomposition for collision-free navigation. Huber \textit{et al}. \cite{huber2019avoidance} adopted contraction-based dynamical systems theory to achieve dynamic obstacle avoidance. Berkane \textit{et al}. \cite{berkane2021navigation} proposed a sensor-based navigation algorithm that employs Nagumo's invariance theorem to guarantee robot safety by projecting the nominal velocity onto safe velocity cones (e.g., Bouligand's tangent cones) when the robot is close to obstacle boundaries. Although most reactive approaches offer closed-form solutions, they typically consider only the nominal robot kinematics, without accounting for disturbances that may arise, for example, from sensor noise, imperfect wheel alignment, and motor or actuator asymmetry. These unmodeled factors inevitably degrade system performance, potentially leading to deviations of the desired trajectory and compromising the robot's safety. Furthermore, the aforementioned methods can only guarantee that WMRs reach the desired goal asymptotically, which is undesirable for time-critical missions. 

There has been extensive research on the trajectory tracking control of WMRs. Zhai and Song \cite{zhai2019adaptive} combined the adaptive and sliding mode control techniques to address the trajectory tracking problem of WMRs. Zheng \emph{et al.} \cite{zheng2024adaptive} presented an adaptive sliding mode control method for trajectory tracking of WMRs, and introduced a barrier function-like control gain to prevent input saturation. Kim and Singh\cite{kim2023energy} proposed a differential-flatness-based feedback controller to achieve the tracking of prescribed point-to-point trajectories. Recently, Zhou \emph{et al.}\cite{zhou2023homogeneity} proposed a homogeneous control strategy to solve the trajectory tracking problem for perturbed unicycle mobile robots, where the convergence rate can be tuned by selecting a proper homogeneity degree. While these methods are effective in addressing stability and steady-state performance, they often do not explicitly consider the critical requirement of maintaining the WMR's actual motion within a predefined safe tube around the reference trajectory. This actually imposes constraints on the tracking errors. Furthermore, most of these methods can only achieve asymptotic position tracking, and they typically lack the capability to preassign the convergence time according to the task requirements.

This paper examines the planning and control problems for prescribed-time collision-free navigation of WMRs in a convex workspace cluttered with static, sufficiently separated, and convex obstacles. By selecting an off-axis point as the virtual control point, the non-holonomic kinematics of WMRs is transformed into a fully-actuated system \cite[Section 11.6.1]{siciliano2009mobile}). Building upon this model, we propose a novel approach, called \textit{InPTC} (Integrated Planning and Tube-Following Control), to achieve collision-free navigation of WMRs within a finite time that can be preassigned according to the task requirement, even in the presence of disturbances. Numerical simulations and experiments illustrate the effectiveness of the proposed InPTC. The main contributions of this paper are twofold:

\begin{enumerate} [1)]
	\item A reactive path planner with prescribed-time convergence is developed based on Bouligand's tangent cones \cite{berkane2021navigation} and time scale transformation \cite{tran2020finite}. This planner yields a continuous vector field that can guide the WMR from almost all initial positions in the free space to reach the goal at a prescribed finite time, while avoiding entering the obstacle regions augmented by safety margins.  
	
	
	
	
	
	\item By leveraging time scale transformation and barrier functions, a prescribed-time tube-following controller is derived for reference trajectory tracking. This controller guarantees that the tracking error converges to a small residual set within a prescribed finite time, whilst the WMR's position remains within a predefined safe tube around the reference trajectory, despite the presence of disturbances. 
\end{enumerate}

The remainder of the paper is organized as follows: In Section \ref{secII}, we describe the kinematics and operating environment of WMRs, and formulate the prescribed-time safe navigation problem. In Section \ref{secIII}, an integrated planning and control framework is proposed to address this problem. The simulation and experimental results are presented in Sections \ref{secIV} and \ref{secV}, respectively. Finally, Section \ref{conclusion} provides concluding remarks and discusses future work.


\textit{Notations:} Throughout the paper, $\mathbb{R}^{n}$ is the $n$-dimensonal Euclidean space, $\mathbb{R}^{m \times n}$ is the vector space of $m\times n$ real matrices, and $\mathbf{I}_{n}$ is a $n\times n$ unit matrix. $|\cdot|$ is the absolute value, and $\|\cdot\|$ denotes either the Euclidean vector norm or the induced matrix norm. The topological interior and boundary of a subset $\mathcal{A}\subset\mathbb{R}^{n}$ are denoted by $\text{int}(\mathcal{A})$ and $\partial\mathcal{A}$, respectively, while the complement of $\mathcal{A}$ in $\mathbb{R}^{n}$ is denoted by $\complement \mathcal{A}$. Given two non-empty subsets $\mathcal{A},\mathcal{B}\subset\mathbb{R}^{n}$, $\text{d}_{\mathcal{A}}(\mathbf{x}):=\inf\{\|\mathbf{x}-\mathbf{y}\|\mid \mathbf{y}\in\mathcal{A}\}$ denotes the distance of a point $\mathbf{x}\in\mathbb{R}^{n}$ to the set $\mathcal{A}$, and $\text{d}(\mathcal{A},\mathcal{B}):=\inf\{\|\mathbf{a}-\mathbf{b}\|\mid \mathbf{a}\in\mathcal{A},~\mathbf{b}\in\mathcal{B}\}$ denotes the distance between $\mathcal{A}$ and $\mathcal{B}$. 

\section{Problem Statement} \label{secII}

\subsection{Kinematics of wheeled mobile robots}

\begin{figure}[hbt!]
	\centering
	\includegraphics[width=5cm]{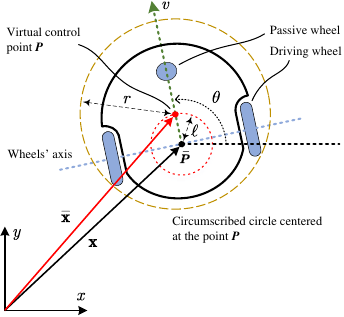}
	\caption{Sketch of the wheeled mobile robot.}
	\label{WMR}
\end{figure}

In this paper, we consider a nonholonomic wheeled mobile robot (WMR) with a control point $\bar{\bm{P}}$
located at the midpoint of the axis connecting the two driving wheels, as depicted in Fig. \ref{WMR}. Let $\bar{\mathbf{x}}:=[\bar{x},\bar{y}]^{\top}\in\mathbb{R}^{2}$ be the position vector of $\bar{\bm{P}}$ in the global coordinate frame, and $\bar{\theta}$ be the WMR's heading angle. The kinematic model of the WMR is expressed as
\begin{equation}
\label{eq1}
\left[
\begin{matrix}
\dot{\bar{x}}\\
\dot{\bar{y}}\\
\dot{\bar{\theta}}
\end{matrix}\right] =
\left[
\begin{matrix}
\cos\bar{\theta} & 0 \\
\sin\bar{\theta} & 0 \\	
0            & 1
\end{matrix}
\right](\mathbf{u}+\mathbf{u}_{d}),
\end{equation}
where $\mathbf{u}:=[v,\omega]^{\top}\in\mathbb{R}^{2}$ is the the control input vector with $v$ and $\omega$ being the linear and angular velocities of the WMR, respectively, and $\mathbf{u}_d\in\mathbb{R}^{2}$ is the matched disturbance that may arise from sensor noise, imperfect wheel alignment, and motor or actuator asymmetry. 


To facilitate the subsequent design and analysis, we select an off-axis point $\bm{P}$ as the virtual control point, which locates at a distance $\ell \neq 0$ away from $\bar{\bm{P}}$ along the longitudinal axis of the WMR \cite{siciliano2009mobile}, as shown in Fig. \ref{WMR}. Denote by $\mathbf{x}:=[x,y]^{\top}\in\mathbb{R}^{2}$ and $\theta$ the position and heading angle of $\bm{P}$, respectively. We have the following change of coordinates:
\begin{equation}
\label{eq2}
\left\lbrace
\begin{aligned}
&x:=\bar{x}+\ell\cos\theta,\\
&y:=\bar{y}+\ell\sin\theta,\\
&\theta:=\bar{\theta}.
\end{aligned}\right.
\end{equation}
In view of \eqref{eq1} and \eqref{eq2}, the kinematics of $\mathbf{x}$ is given by
\begin{equation}
\label{kine_dx}
\dot{\mathbf{x}}=\mathbf{R}(\theta)(\mathbf{u}+\mathbf{u}_d),
\end{equation}
where $\mathbf{R}(\theta):=[\cos\theta, -\ell\sin\theta; \sin\theta, \ell\cos\theta]$ is full rank. This enables us to freely steer the WMR's position regardless of nonholonomic constraints. By setting $|\ell| \leq 1$, it follows that $\|\mathbf{R}(\theta)\|=\max\{1,|\ell|\}=1$.

\begin{assumption} \label{A1}
	The disturbance $\mathbf{u}_d$ is bounded by an unknown constant $d=\sup_{t\geq0}\|\mathbf{u}_d(t)\|$, that is, $\|\mathbf{u}_d\|\leq d$.  
\end{assumption}

\begin{remark}
The kinematics given by \eqref{eq1} represents a general kinematic model for unicycle mobile robots, such as differential drive robots and synchro drive robots. Additionally, following the input/output linearization outlined in \cite[Section 11.6.1]{siciliano2009mobile}, the nonholonomic kinematics \eqref{eq1} is transformed into a full-actuated system \eqref{kine_dx}, via a change of coordinates given by \eqref{eq2}. This transformed model can be rewritten in the form of a single integrator system, which has the same structure as the kinematics of omnidirectional robots with holonomic constraints. Consequently, the proposed method building upon \eqref{kine_dx} is applicable for a broad class of WMRs.
\end{remark}

\subsection{Operating environment} \label{secII-B}

Consider a WMR operating inside a closed compact convex workspace $\mathcal{W}^{\ast}\subset\mathbb{R}^{2}$, punctured by a set of $n$ static obstacles $\mathcal{O}_{i}^{\ast}$, $i\in\mathbb{I}:=\{1,2,...,n\}$, which are represented by open balls with centers $\mathbf{c}_{i}\in\mathbb{R}^{2}$ and radii $r_{i}>0$. A circumscribed circle centered at $\bm{P}$ with radius $r>0$ is constructed, which is the smallest circle enclosing the WMR (see Fig. \ref{WMR}). To ensure that the WMR can navigate freely between any of the obstacles in $\mathcal{W}^{\ast}$, we make the following assumption \cite{berkane2021navigation,arslan2019sensor}: 

\begin{assumption} \label{A2}
	The $n$ obstacles are separated from each other by a clearance of at least
	\begin{equation}
	\label{eq5}
	{\rm d}(\mathcal{O}_{i}^{\ast},\mathcal{O}_{j}^{\ast})>2(r+h),~\forall i,j\in \mathbb{I},~i \neq j,
	\end{equation}
	and from the boundary of the workspace $\mathcal{W}^{\ast}$ as
	\begin{equation}
	\label{eq6}
	{\rm d}(\mathcal{O}_{i}^{\ast},\partial\mathcal{W}^{\ast})>2r+h,~\forall i\in \mathbb{I},
	\end{equation} 
	where $h>0$ is a constant.
\end{assumption}

\begin{figure}[hbt!]
	\centering
	\includegraphics[width=8.5cm]{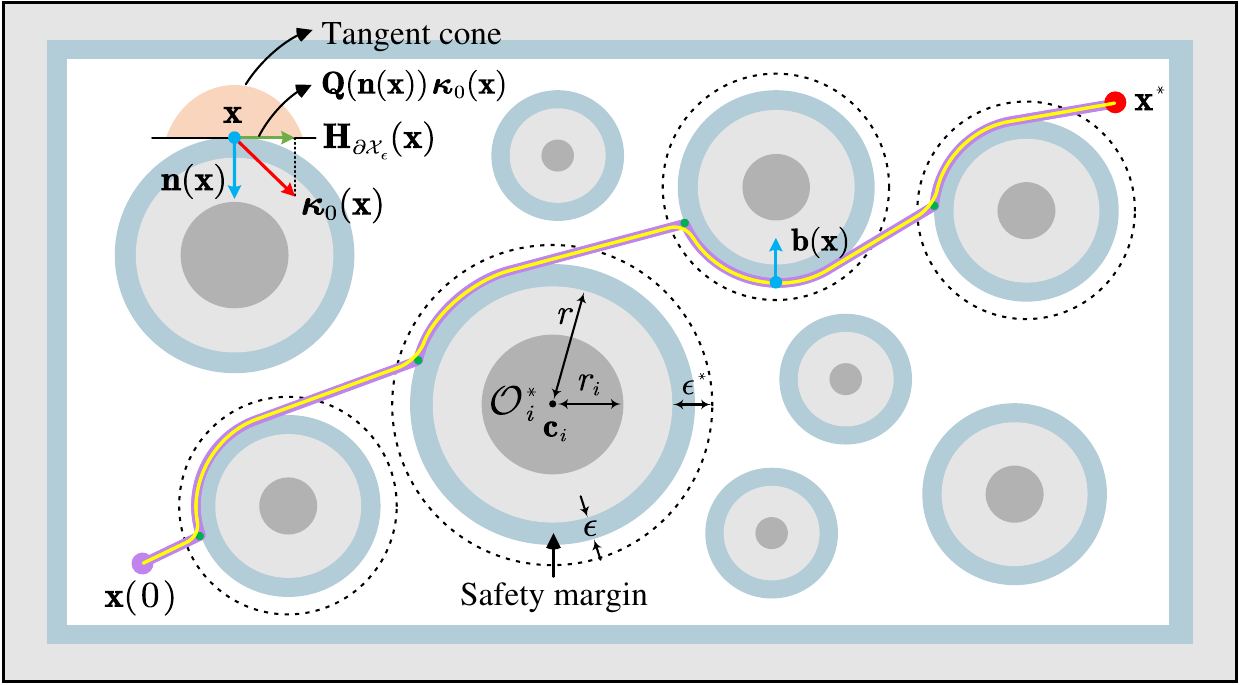}
	\caption{Schematic diagram of the operating environment, where the dark gray balls denote the actual obstacles $\mathcal{O}_{i}^{\ast}$, the light gray regions denote the augmented obstacle regions $\mathcal{O}_{i}$, and the light blue regions denote the safety margin. In addition, the dashed circles are the influence regions of obstacles, whereas the pink and yellow lines denote the discontinuous trajectory and its continuous counterpart, respectively.}
	\label{schematic_diagram}
\end{figure}

For ease of design, the WMR is considered as a point (i.e., the off-axis point $\bm{P}$), by transferring the volume of the circumscribed circle to the other workspace entities. For the point $\bm{P}$, the workspace is $\mathcal{W}:=\{\mathbf{x}\in\mathbb{R}^{2}\mid\text{d}_{\complement \mathcal{W}^{\ast}}(\mathbf{x})\geq r\}$, and the obstacle regions are $\mathbb{R}^{2}$: $\mathcal{O}_{i}:=\{\mathbf{x}\in\mathbb{R}^{2}\mid\beta_i(\mathbf{x})<0\},~i\in\mathbb{I}$, where $\beta_i(\mathbf{x}):=\|\mathbf{x}-\mathbf{c}_{i}\|-(r+r_{i})$. Thus, the free space of $\bm{P}$ is given by the closed set $\mathcal{X}:=\mathcal{W}\setminus \mathcal{O}$ with $ \mathcal{O}:=\cup_{i=1}^{n}\mathcal{O}_{i}$. To enhance the safety, a safety margin of size $0<\epsilon<h$ is introduced (see the light blue regions in Fig. \ref{schematic_diagram}), which results in an eroded workspace $\mathcal{W}^{\epsilon}:=\{\mathbf{x}\in\mathbb{R}^{2}\mid\text{d}_{\complement \mathcal{W}^{\ast}}(\mathbf{x})\geq r+\epsilon\}$ and $n$ augmented obstacle regions $\mathcal{O}_{i}^{\epsilon}:=\{\mathbf{x}\in\mathbb{R}^{2}\mid\beta_i(\mathbf{x})<\epsilon\}$, $i\in\mathbb{I}$. Then, the free space of $\bm{P}$ reduces to $\mathcal{X}_{\epsilon}:=\mathcal{W}^{\epsilon}\setminus\mathcal{O}^{\epsilon}$ with $ \mathcal{O}^{\epsilon}:=\cup_{i=1}^{n}\mathcal{O}_{i}^{\epsilon}$.

\subsection{Problem formulation}

The robot navigation problem is formulated as follows:

\begin{problem} \label{P1}
	Consider the WMR kinematics described by \eqref{kine_dx} under Assumptions \ref{A1} and \ref{A2}. The objective is to derive a control law $\mathbf{u}$ that derives the virtual control point $\bm{P}$ from an initial position $\mathbf{x}(0)\in\mathcal{X}_{\epsilon}$ to a designated goal $\mathbf{x}^{\ast}\in\text{int}(\mathcal{X}_{\epsilon})$, without colliding with any obstacles along the route.
\end{problem}
	
\section{Main Results} \label{secIII}

\begin{figure}[hbt!]
	\centering
	\includegraphics[width=8.5cm]{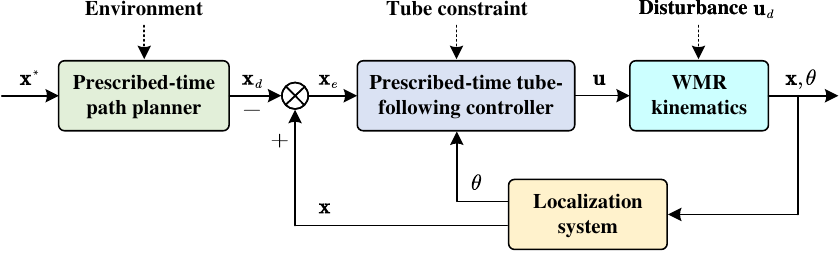}
	\caption{Block diagram of the proposed InPTC scheme.}
	\label{block_diagram}
\end{figure}

In this section, an approach, called InPTC, is proposed to solve Problem \ref{P1}. The block diagram of this scheme is shown in Fig. \ref{block_diagram}, where the planner module generates a collision-free reference trajectory $\mathbf{x}_{d}$ with prescribed-time convergence. In the control module, a tube-following controller is derived to achieve prescribed-time trajectory tracking within a safe tube around $\mathbf{x}_{d}$ (with radius less than the safety margin). Under this framework, $\mathbf{x}$ will converge to the goal $\mathbf{x}^\ast$ within a prescribed time, without colliding with any obstacles along the way, thus achieving prescribed-time safe robot navigation. 

\subsection{Preliminaries}

\begin{definition}[Bouligand's tangent cone  \cite{bouligand1932introduction}] \label{D1}
	Given a closed set $\mathcal{F}\in\mathbb{R}^{n}$, the tangent cone to $\mathcal{F}$ at a point $\mathbf{x}\in\mathbb{R}^{n}$ is the subset of $\mathbb{R}^{n}$ defined by
	\begin{equation}
	\nonumber
	\mathbf{T}_{\mathcal{F}}(\mathbf{x}):=\left\lbrace \mathbf{z}\in\mathbb{R}^{n} \bigg|\mathop{\lim \inf}\limits_{\tau\to 0^{+}}\dfrac{{\rm d}_{\mathcal{F}}(\mathbf{x}+\tau \mathbf{z})}{\tau}=0
	\right\rbrace. 
	\end{equation}
\end{definition}

The tangent cone $\mathbf{T}_{\mathcal{F}}(\mathbf{x})$ is a set that contains all the vectors pointing from $\mathbf{x}$ inside or tangent to $\mathcal{F}$, while for $\mathbf{x}\notin \mathcal{F}$, $\mathbf{T}_{\mathcal{F}}(\mathbf{x})=\emptyset$. Since for all $\mathbf{x}\in \text{int}(\mathcal{F})$, we have $\mathbf{T}_{\mathcal{F}}(\mathbf{x})=\mathbb{R}^{n}$, the tangent cone $\mathbf{T}_{\mathcal{F}}(\mathbf{x})$ is non-trivial only on the boundary $\partial\mathcal{F}$. Next, we recall the Nagumo's invariance theorem.

\begin{theorem}	[Nagumo 1942 \cite{nagumo1942lage}] \label{Th1} 
	Consider the system $\dot{\mathbf{x}}(t)=\mathbf{f}(\mathbf{x}(t))$, which admits a unique solution in forward time for each initial condition $\mathbf{x}(0)$ in an open set $\mathcal{O}$. The closed set $\mathcal{F}\subset\mathcal{O}$ is forward invariant iff $\mathbf{f}(\mathbf{x})\in\mathbf{T}_{\mathcal{F}}(\mathbf{x})$, $\forall\mathbf{x}\in\mathcal{F}$.
\end{theorem}



\begin{definition} [Time scale transformation function, TSTF \cite{tran2020finite}] \label{D3}
	A smooth function $\eta:[0,\infty)\to[0,T)$ is called a TSTF if 1) it is strictly increasing; 2) it is s.t.  $\eta(0)=0$ and $\eta^\prime(0) =1$; 3) it is s.t. $\lim_{s \to \infty}\eta(s)=T$ and $\lim_{s\to \infty}\eta^\prime(s)=0$.
\end{definition}

A TSTF squeezes the infinite-time interval $s\in[0,\infty)$ into a prescribed finite time interval $t:=\eta(s)\in[0,T)$, which plays a key role for achieving prescribed-time planning and control. In this work, $\eta(s)$ is of the form
\begin{equation}
\label{eta}
\eta(s)=T(1 - e^{-\frac{s}{T}}).
\end{equation}


\subsection{Prescribed-time path planning} \label{secIII-B}


We neglect the disturbance $\mathbf{u}_d$ in \eqref{kine_dx} and consider a control law $\mathbf{u}=\mathbf{R}^{-1}(\theta)\bm{\tau}$, where $\bm{\tau}\in\mathbb{R}^{2}$ is a virtual control law to be designed. Then, substituting it into \eqref{kine_dx}, one gets
\begin{equation}
\label{dx}
\dot{\mathbf{x}}=\bm{\tau}.
\end{equation}
A tangent cone based control policy (vector field) $\bm{\tau}=\mathbf{h}(\mathbf{x})$ is designed to achieve collision-free path planning. According to Theorem \ref{Th1}, we consider a nearest point problem
\begin{equation}
\label{eq12}
\begin{split}
\min\limits_{\mathbf{h}}&~\|\mathbf{h}(\mathbf{x})-\bm{\kappa}_{0}(\mathbf{x})\|\\
\text{s.t.}&~ \mathbf{h}(\mathbf{x})\in\mathbf{T}_{\mathcal{{X}}_{\epsilon}}(\mathbf{x}),~\forall \mathbf{x}\in\mathcal{X}_{\epsilon},
\end{split}
\end{equation} 
where $\bm{\kappa}_{0}(\mathbf{x})$ is a nominal control law for motion-to-goal and is designed here as
\begin{equation}
\label{kappa_0}
\bm{\kappa}_{0}(\mathbf{x})=-k_0(\mathbf{x}-\mathbf{x}^{\ast}),
\end{equation}
with $k_0>0$ being a constant.

The robot workspace $\mathcal{W}^{\ast}$ is a convex set, so does $\mathcal{W}^{\epsilon}$, which suggests that for all $\mathbf{x}\in\partial\mathcal{W}^{\epsilon}$, $\bm{\kappa}_{0}(\mathbf{x})$ points inside the free space $\mathcal{X}_{\epsilon}$ (i.e., $\bm{\kappa}_{0}(\mathbf{x})\in\mathbf{T}_{\mathcal{{X}}_{\epsilon}}(\mathbf{x})$) and thus is a solution to \eqref{eq12}. Moreover, for all $\mathbf{x}\in \text{int}(\mathcal{X}_{\epsilon})$, $\bm{\kappa}_{0}(\mathbf{x})$ is also the solution to \eqref{eq12}. Next we check the obstacle boundary points $\mathbf{x}\in\partial\mathcal{O}^{\epsilon}$. As $\partial\mathcal{O}^{\epsilon}$ is a smooth hypersurface of $\mathbb{R}^{2}$ and is orientable, there exists a continuously differentiable map (known as Gauss map \cite{lee2022manifolds}) $\mathbf{n}:\partial\mathcal{O}^{\epsilon}\to\mathbb{S}^{1}$ such that for all $\mathbf{x}\in\partial\mathcal{O}^{\epsilon}$, $\mathbf{n}(\mathbf{x})$ is the outward unit normal vector to $\partial\mathcal{O}^{\epsilon}$. As clearly seen in Fig. \ref{schematic_diagram}, the tangent cone at any $\mathbf{x}\in\partial\mathcal{O}^{\epsilon}$ is a half-space $\mathbf{T}_{\mathcal{X}_{\epsilon}}(\mathbf{x}):=\{\mathbf{y}\in\mathbb{R}^{2}\mid(\mathbf{y}-\mathbf{x})^{\top}\mathbf{n}(\mathbf{x})\leq0\}$, which is a convex function. Thus, \eqref{eq12} has a unique solution. In summary, if $\mathbf{x}\in\mathcal{X}_{\epsilon}\setminus\partial\mathcal{O}^{\epsilon}$ or $\mathbf{x}\in\partial\mathcal{O}^{\epsilon}\wedge\bm{\kappa}_{0}^{\top}(\mathbf{x})\mathbf{n}(\mathbf{x})\leq0$, then $\bm{\kappa}_{0}(\mathbf{x})$ is a solution to \eqref{eq12}. While if $\mathbf{x}\in\mathcal{O}^{\epsilon}\wedge\bm{\kappa}_{0}^{\top}(\mathbf{x})\mathbf{n}(\mathbf{x})>0$, the closest point of \eqref{eq12} is obtained by the orthogonal projection $\mathbf{Q}(\mathbf{n}(\mathbf{x})):=\mathbf{I}_{2}-\mathbf{n}(\mathbf{x})\mathbf{n}^{\top}(\mathbf{x})$ onto the tangent hyperplane of $\partial\mathcal{X}_{\epsilon}$, defined by 
$\mathbf{H}_{\partial\mathcal{X}_{\epsilon}}(\mathbf{x}):=\{\mathbf{y}\in\mathbb{R}^{2}\mid(\mathbf{y}-\mathbf{x})^{\top}\mathbf{n}(\mathbf{x})=0\}$. Then, a general solution to \eqref{eq12} is as follows: $\mathbf{h}(\mathbf{x})=\bm{\kappa}_{0}(\mathbf{x})$ if $\text{d}_{\mathcal{O}}(\mathbf{x})>\epsilon$ or $\bm{\kappa}_{0}^{\top}(\mathbf{x})\mathbf{n}(\mathbf{x})\leq0$, and $\mathbf{h}(\mathbf{x})=\mathbf{Q}(\mathbf{n}(\mathbf{x}))
\bm{\kappa}_{0}(\mathbf{x})$ if $\text{d}_{\mathcal{O}}(\mathbf{x})=\epsilon \wedge \bm{\kappa}_{0}^{\top}(\mathbf{x})\mathbf{n}(\mathbf{x})>0$.
The resulting vector field is discontinuous at some boundary points $\mathbf{x}\in\partial\mathcal{O}^{\epsilon}$ like the green points in Fig. \ref{schematic_diagram}. To address this issue, a continuous control law is proposed in the sequel. Following the line of \cite{berkane2021navigation}, we specify an influence region for each obstacle (marked by the dashed line in Fig. \ref{schematic_diagram}), which is defined as $\{\mathbf{x}\in\mathbb{R}^{2}\mid\text{d}_{\mathcal{O}_{i}}(\mathbf{x})\leq\epsilon^{\ast}\}$, where $\epsilon^{\ast}\in(\epsilon,h]$ and $i\in\mathbb{I}$. Obstacle avoidance is activated only when the position $\mathbf{x}$ of $\bm{P}$ enters the influence regions of obstacles. To proceed, a bearing vector is defined as
\begin{equation}
\label{b}
\mathbf{b}(\mathbf{x}):=\dfrac{\mathbf{P}_{\partial\mathcal{O}}(\mathbf{x})-\mathbf{x}}{\|\mathbf{P}_{\partial\mathcal{O}}(\mathbf{x})-\mathbf{x}\|}=\dfrac{\mathbf{P}_{\partial\mathcal{O}}(\mathbf{x})-\mathbf{x}}{\text{d}_{\mathcal{O}}(\mathbf{x})},
\end{equation}
where $\mathbf{P}_{\partial\mathcal{O}}(\mathbf{x}):=\{\mathbf{y}\in\partial\mathcal{O}\mid\|\mathbf{y}-\mathbf{x}\|=\text{d}_{\mathcal{O}}(\mathbf{x})\}$ is a set-valued map. Since $\text{d}(\mathcal{O}_{i}^{\ast},\mathcal{O}_{j}^{\ast})>2(r+h)$, $\forall i,j\in \mathbb{I}$, $i \neq j$ (referring to Assumption \ref{A2}) and $\epsilon^{\ast}\leq h$, there can be only one obstacle $\mathcal{O}_{i}$ such that $\text{d}_{\mathcal{O}}(\mathbf{x})=\text{d}_{\mathcal{O}_{i}}(\mathbf{x})\leq\epsilon^{\ast}$. With this in mind, $\mathbf{b}(\mathbf{x})$ in \eqref{b} can be computed by $\mathbf{b}(\mathbf{x})=(\mathbf{c}_{i}-\mathbf{x})/\|\mathbf{c}_{i}-\mathbf{x}\|$. Note that when $\mathbf{x}\in\partial\mathcal{O}_{\epsilon}$, $\mathbf{b}(\mathbf{x})$ is equivalent to the Gauss map $\mathbf{n}(\mathbf{x})$. Now the control law is modified as 
\begin{equation}
\label{h}
\mathbf{h}(\mathbf{x})=\left\lbrace
\begin{split}
&\bm{\kappa}_{0}(\mathbf{x}),&\text{if}~&\text{d}_{\mathcal{O}}(\mathbf{x})>\epsilon^{\ast}~\text{or}\\
&&&\bm{\kappa}_{0}^{\top}(\mathbf{x})\mathbf{b}(\mathbf{x})\leq0, \\
&\mathbf{\Pi}(\mathbf{x})
\bm{\kappa}_{0}(\mathbf{x}),&\text{if}~&\text{d}_{\mathcal{O}}(\mathbf{x})\leq\epsilon^{\ast}~\text{and}\\
&&&\bm{\kappa}_{0}^{\top}(\mathbf{x})\mathbf{b}(\mathbf{x})>0,
\end{split}\right.
\end{equation}
with
\begin{equation}
\label{Pi_x}
\mathbf{\Pi}(\mathbf{x}):=\mathbf{I}_{2}-\phi(\text{d}_{\mathcal{O}}(\mathbf{x}))\mathbf{b}(\mathbf{x})\mathbf{b}^{\top}(\mathbf{x}),
\end{equation}
where $\phi(\text{d}_{\mathcal{O}}(\mathbf{x}))\in\mathbb{R}$ is a $C^{1}$ bump function that smoothly transitions from $1$ to $0$ on the interval $\text{d}_{\mathcal{O}}(\mathbf{x})\in[\epsilon,\epsilon^{\ast}]$. A simple choice of $\phi(\text{d}_{\mathcal{O}}(\mathbf{x}))$ is 
\begin{equation}
\label{bump}
\phi(\text{d}_{\mathcal{O}}(\mathbf{x}))=\left\lbrace
\begin{aligned}
&1, &\text{if}~&\text{d}_{\mathcal{O}}(\mathbf{x})\leq\epsilon,\\
&\lambda(\text{d}_{\mathcal{O}}(\mathbf{x})), &\text{if}~&\epsilon<\text{d}_{\mathcal{O}}(\mathbf{x})<\epsilon^{\ast}, \\
&0,&\text{if}~& \text{d}_{\mathcal{O}}(\mathbf{x})\geq\epsilon^{\ast},
\end{aligned}
\right. 
\end{equation}
where $\lambda(\text{d}_{\mathcal{O}}(\mathbf{x}))=0.5[1-\cos(\pi(\epsilon^{\ast}-\text{d}_{\mathcal{O}}(\mathbf{x}))/(\epsilon^{\ast}-\epsilon))]$.
It is clear that the modified control law \eqref{h} is continuous and piecewise continuously differentiable on the domain $\mathcal{X}$. 

\begin{theorem} \label{Th2}
	Consider the kinematics defined by \eqref{dx}. If $\mathbf{x}^{\ast}\in\text{int}(\mathcal{X}_{\epsilon})$ and the obstacles $\mathcal{O}_{i}^{\ast}$, $i\in\mathbb{I}$ satisfying Assumption \ref{A2}, then the continuous control law \eqref{h} with $\bm{\kappa}_{0}(\mathbf{x})$ given by \eqref{kappa_0} can guarantee that:
	\begin{enumerate}
		\item The free space $\mathcal{X}_{\epsilon}$ is forward invariant.
		\item For any $\mathbf{x}(0)\in\mathcal{X}_{\epsilon}$, the solution of \eqref{dx} admits a unique solution in forward time, which asymptotically converges to the set $\mathcal{C}:=\{\mathbf{x}^{\ast}\}\cup_{i=1}^{n}\{\mathbf{s}_{i}\}$, where the stationary point $\mathbf{s}_{i}$ that satisfies ${\rm d}_{\mathcal{O}_{i}}(\mathbf{s}_{i})=\epsilon \wedge \bm{\kappa}_{0}^{\top}(\mathbf{s}_{i})\mathbf{b}(\mathbf{s}_{i})/\|\bm{\kappa}_{0}(\mathbf{s}_{i})\|=1$ is locally unstable.
		\item The equilibrium point $\mathbf{x}=\mathbf{x}^{\ast}$ is almost globally asymptotically stable and locally exponentially stable.  
	\end{enumerate} 
\end{theorem}

\textit{Proof}. The proof is relegated to Appendix. $\hfill \blacksquare$


To further achieve prescribed-time path planning, the TSTF $t:=\eta(s)$ defined in \eqref{eta} is introduced to squeeze and map the collision-free trajectory of \eqref{dx} from an infinite time interval $s\in[0,\infty)$ to a prescribed finite time interval $t\in[0,T)$. Let $\eta^\prime(s):=d\eta(s)/ds$ and define a prescribed-time gain function (PTGF) $\alpha:\mathbb{R}_{\geq0}\to\mathbb{R}_{>0}$ as
\begin{equation}
\label{alpha}
\alpha(t)=\alpha(\eta(s)):=\dfrac{1}{\eta^\prime(s)}=\dfrac{T}{T-t},~ t\in[0,T),
\end{equation}
showing that $\alpha(t)>0$ is continuously differentiable on $t\in[0,T)$ and satisfies $\alpha(0)=1$ and $\lim_{t\to T}\alpha(t)=\infty$. Here the prescribed-time control law is designed as $\bm{\tau}=\mathbf{h}^\ast(\mathbf{x},t)$, where $\mathbf{h}^\ast$ is given by 
\begin{equation}
\label{h_ast}
\mathbf{h}^\ast(\mathbf{x},t)=\left\lbrace
\begin{split}
&\alpha(t)\mathbf{h}(\mathbf{x}), &t&\in[0,T), \\
&\mathbf{h}(\mathbf{x}), &t&\in[T,\infty),
\end{split}\right.
\end{equation}
with $\mathbf{h}(\mathbf{x})$ and $\alpha(t)$ defined in \eqref{h} and \eqref{alpha}, respectively. 

\begin{theorem} \label{Th3}
    Consider the kinematics given by \eqref{dx} with initial conditions satisfying $\mathbf{x}(0)\in\mathcal{X}_{\epsilon}$. Then, the closed-loop system $\dot{\mathbf{x}}=\mathbf{h}^\ast(\mathbf{x},t)$ converge to the set $\mathcal{C}$ at $t=T$, while avoiding obstacle regions $\mathcal{O}_i^\epsilon$, $i\in\mathbb{I}$. The system under $\bm{\tau}=\mathbf{h}^\ast(\mathbf{x},t)$ follows the same path as the closed-loop system $\dot{\mathbf{x}}=\mathbf{h}(\mathbf{x})$, but goes faster within the time window $[0,T)$.
\end{theorem}

\textit{Proof:} The proof is conducted over two time intervals.

1) Consider the interval $t\in[0,T)$. The closed-loop system is written as
\begin{equation}
\label{dx_1}
\dot{\mathbf{x}}(t)=\alpha(t)\mathbf{h}(\mathbf{x}(t)).
\end{equation}
Let $\bar{\mathbf{x}}(s):=\mathbf{x}(t)=\mathbf{x}(\eta(s))$. Then, the system \eqref{dx_1} is rewritten in the stretched infinite-time interval $s\in[0,\infty)$ as
\begin{equation}
\label{dx_pri}
\bar{\mathbf{x}}^\prime(s):=\dfrac{d\bar{\mathbf{x}}(s)}{ds}=\dfrac{d\mathbf{x}(\eta(s))}{d\eta(s)}\cdot\dfrac{d\eta(s)}{ds}.
\end{equation}
In view of \ref{alpha} and \eqref{dx_1}, \ref{dx_pri} becomes
\begin{equation}
\label{dx_pri_bar}
\bar{\mathbf{x}}^\prime(s)=\mathbf{h}(\bar{\mathbf{x}}(s)).
\end{equation}
The initial conditions satisfy $\bar{\mathbf{x}}^\prime(0)=\dot{\mathbf{x}}(0)$ and $\bar{\mathbf{x}}(0)=\mathbf{x}(0)$, where the facts that $\eta(0)=0$ and $\alpha(0)=1$ have been used. As \eqref{dx_pri_bar} has the same solution as the closed-loop system $\dot{\mathbf{x}}(t)=\mathbf{h}(\mathbf{x}(t))$, it follows from Theorem \ref{Th2} that $\bar{\mathbf{x}}(s)$ asymptotically converges to the set $\mathcal{C}$. This implies that $\mathbf{x}(t)$ converges to the set $\mathcal{C}$ at the prescribed time $T$, due to $\mathbf{x}(t)=\bar{\mathbf{x}}(s)$ and $t\to T$ as $s\to\infty$. Additionally, Theorem \ref{Th2} ensures that $\bar{\mathbf{x}}(s)$ is collision-free on $t\in[0,\infty)$. As a result, $\mathbf{x}(t)$ is also  collision-free on $t\in[0,T)$.

2) Let us consider the interval $t\in[T,\infty)$, on which the closed-loop system becomes $\dot{\mathbf{x}}(t)=\mathbf{h}(\mathbf{x}(t))$. It can be verified that $\mathbf{h}(\mathbf{x}(t))=\mathbf{0}$ for all $t\geq T$, no matter $\mathbf{x}(t)\to\mathbf{x}^{\ast}$ or $\mathbf{x}(t)\to\bm{s}_i$ as $t\to T$, indicating that the system states remain unchanged. This completes the proof. $\hfill \blacksquare$ 	

\begin{remark}	
	The proposed prescribed-time control policy \eqref{h_ast} is discontinuous at $t=T$ and cannot be practically implemented due to the unboundedness of $\alpha(t)$ as $t\to T$. To remedy this, a continuous version is given as
	\begin{equation}
	\label{h_ast1}
	\tau=\mathbf{h}^\ast(\mathbf{x},t)=\alpha^\ast(t)\mathbf{h}(\mathbf{x}),
	\end{equation}
	with the PTGF given by
	\begin{equation}
	\label{alpha_ast}
	\alpha^\ast(t):=\left\lbrace
	\begin{aligned}
	&\alpha(t),   &~&t\in[0,T^\ast), \\
	&\alpha(T^\ast),&~&t\in[T^\ast,\infty),
	\end{aligned}\right.
	\end{equation}
	where $T^\ast := T-\varsigma>0$, and $0<\varsigma\ll T$ is a sufficiently small constant. The resulting continuous vector field can guide $\mathbf{x}$ to a small neighborhood of the set $\mathcal{C}$ at $t=T^\ast$. Since $\alpha^\ast(t)\equiv T/\varsigma>0$ on $[T^\ast,\infty)$, $\mathbf{x}(t)$ will asymptotically converge to the set $\mathcal{C}$ after $t=T^\ast$. It is clear from \eqref{kappa_0} that choosing a larger $k_0$ will increase the convergence rate of the planner during the transient-state phase. However, this comes at the expense of increased reference velocity.
\end{remark}


\begin{remark}
	Although the prescribed-time path planner \eqref{h_ast1} is partially inspired by \cite{berkane2021navigation}, there remain substantial differences in the following two aspects:
	\begin{enumerate}
		\item A $C^1$ bump function $\phi(\text{d}_{\mathcal{O}}(\mathbf{x}))$ defined in \eqref{bump} is introduced into the projection function $\mathbf{\Pi}(\mathbf{x})$, which makes the resulting vector filed more smoother;
		\item By incorporating a PTGF $\alpha(t)$ (generated from the TSTF defined in \eqref{eta}) into the planning algorithm developed in \cite{berkane2021navigation}, the proposed planner \eqref{h_ast1} achieves prescribed-time collision-free path planning.
	\end{enumerate}
    Moreover, the planning algorithm offers an analytical solution without involving any numerical optimization solving, rendering it computationally efficient. 
\end{remark}

\begin{remark}
    Although the path planner in \eqref{h_ast1} is proposed for environments with circular obstacles, it can be easily extended to deal with convex polygonal obstacles, such as rectangle, trapezoid, and triangle. For such obstacles, we can define the augmented obstacle regions $\mathcal{O}_i:=\mathcal{O}_i^\ast\cup\{\mathbf{x}\in\mathbb{R}^{2}\mid{\rm d}_{\mathcal{O}_i^\ast}(\mathbf{x})<r\}$ and $\mathcal{O}_i^\epsilon:=\mathcal{O}_i^\ast\cup\{\mathbf{x}\in\mathbb{R}^{2}\mid{\rm d}_{\mathcal{O}_i^\ast}(\mathbf{x})<r+\epsilon\}$. The distance function ensures that the boundaries of $\mathcal{O}_i$ and $\mathcal{O}_i^\epsilon$ are smooth, as shown in Fig. \ref{fig_SGF}. In practical implementation, the distance function ${\rm d}_{\mathcal{O}}(\mathbf{x})$ and the bearing vector $\mathbf{b}(\mathbf{x})$ in $\mathbf{h}(\mathbf{x})$ can be obtained locally using on-board LiDARs.
   
    \begin{figure}[htb!]
    	\centering
    	\includegraphics[width=8cm]{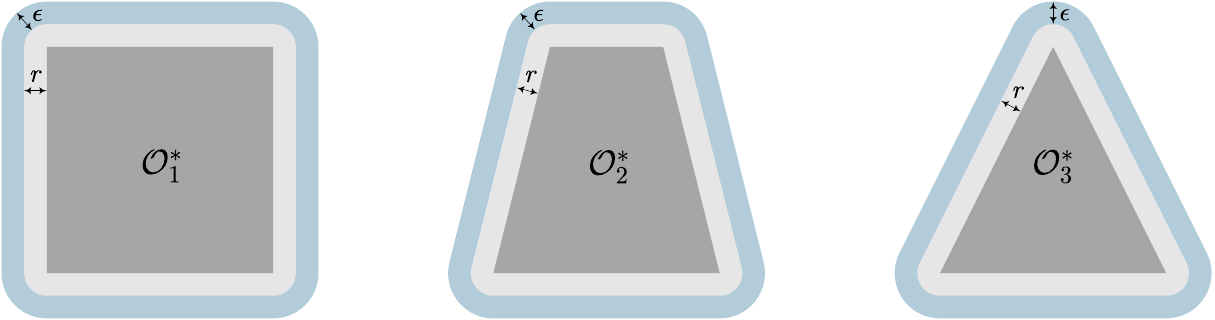}
    	\caption{Augmented regions of convex polygonal obstacles.}
    	\label{fig_SGF}
    \end{figure}
\end{remark}

\subsection{Prescribed-time tube-following control}

\begin{figure}[htb!]
	\centering
	\includegraphics[width=4cm]{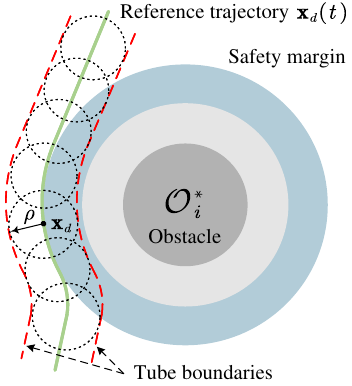}
	\caption{Illustration of the predefined safe tube.}
	\label{fig_tube}
\end{figure}

To avoid conflicts of notations, the reference trajectory is denoted as $\mathbf{x}_{d}$, which is governed by $\dot{\mathbf{x}}_{d}=\bm{\tau}_{d}$, where $\bm{\tau}_{d}$ is given by \eqref{h_ast1} but with $\mathbf{x}$ replaced by $\mathbf{x}_{d}$. Define the position tracking error as $\mathbf{x}_{e}:=\mathbf{x}-\mathbf{x}_{d}$. It follows from \eqref{kine_dx} that 
\begin{equation}
\label{dx_e}
\dot{\mathbf{x}}_e = \mathbf{R}(\theta)(\mathbf{u}+\mathbf{u}_d)-\bm{\tau}_{d}.
\end{equation}
To ensure the tracking safety, we impose the following ``tube'' constraint on $\mathbf{x}_e$
\begin{equation}
\label{tube}
\|\mathbf{x}_{e}(t)\|^2<\rho^2,
\end{equation}  
where $\rho>0$ is a user-defined constant. At each time instance, the actual position $\mathbf{x}$ of the virtual control point $\bm{P}$ is allowed to stay within a 1-sphere of radius $\rho$ and center $\mathbf{x}_{d}(t)$. Unifying all such spheres along $t$ forms a tube around $\mathbf{x}_{d}(t)$, as shown in Fig. \ref{fig_tube}. Since a safety margin is considered in the planning module, if the tube radius is taken no larger than the size $\epsilon$ of the safety margin and \eqref{tube} holds for all $t\geq0$, then the actual trajectory $\mathbf{x}(t)$ remains within the safe tube without colliding with any obstacles, that is, $\mathbf{x}(t)\in\mathcal{X}$, $\forall t\geq0$. 

In the following, a tube-following controller is developed to achieve prescribed-time trajectory tracking, while ensuring that $\mathbf{x}(t)$ evolves within the tube defined by \eqref{tube}. To this end, we define a transformed error
\begin{equation}
\label{xi}
\xi(t):=\dfrac{\|\mathbf{x}_{e}(t)\|^2}{\rho^2}.
\end{equation}
Taking the time derivative of \eqref{xi}, whilst using \eqref{kine_dx} and \eqref{dx_e}, one can easily get
\begin{equation}
\label{dxi}
\dot{\xi}=\dfrac{2}{\rho^2}\mathbf{x}_{e}^{\top}[\mathbf{R}(\theta)(\mathbf{u}+\mathbf{u}_d)-\bm{\tau}_{d}].
\end{equation}
Inspecting \eqref{xi} reveals that $0\leq\xi(t)<1$ is equivalent to \eqref{tube}, and $\xi(t)=0$ only when $\mathbf{x}_{e}(t)=\mathbf{0}$. Therefore, the prescribed-time tube-following control problem boils down to achieving $\lim_{t\to T_f}\xi(t)=0$, while ensuring $0\leq\xi(t)<1$, $\forall t\geq0$, via a properly-designed controller. 

To achieve trajectory tracking within a prescribed finite time $T_f>0$, as per Definition \ref{D3}, we introduce a TSTF $t:=\eta_f(s)=T_f(1 - e^{-s/T_f})$, which generates the following PTGF:
\begin{equation}
\label{alpha_f}
\alpha_f(t)=\alpha_f(\eta_f(s)):=\dfrac{1}{\eta_f^\prime(s)}=\dfrac{T_f}{T_f-t},~ t\in[0,T_f).
\end{equation}
Let us define $\mathbf{z}:=\mathbf{x}_{e}/(\rho^2(1-\xi))$ and design the prescribed-time control law as
\begin{equation}
\label{u}
\mathbf{u}=\alpha_f^\ast(t)\mathbf{u}_n+\mathbf{R}^{-1}(\theta)(1-\alpha_f^\ast(t))\bm{\tau}_d,
\end{equation}
with
\begin{equation}
\label{u_n}
\mathbf{u}_n=\mathbf{R}^{-1}(\theta)\left(-k_1\mathbf{x}_{e}-\dfrac{k_2}{\alpha_f^\ast(t)}\mathbf{z}+\bm{\tau}_{d}\right), 
\end{equation}
where $k_1,k_2>0$ are constant gains, and $\alpha_f^\ast(t)$ is a continuous PTGF akin to $\alpha^\ast(t)$ in \eqref{alpha_ast}, given by
\begin{equation}
\alpha_f^\ast(t):=\left\lbrace
\begin{aligned}
&\alpha_f(t),     &~&t\in[0,T_f^\ast), \\
&\alpha_f(T_f^\ast),&~&t\in[T_f^\ast,\infty),
\end{aligned}\right.
\end{equation}
with $T_f^\ast:=T_f-\varsigma_f>0$ and $0<\varsigma_f\ll T_f$ being constants.

\begin{theorem} \label{Th4}
Consider the WMR  kinematics \eqref{kine_dx} under Assumption \ref{A1}. If the initial position $\mathbf{x}(0)$ satisfies \eqref{tube} and the gain $k_1$ is chosen such that $k_1 T_f\gg1$ holds, then the control law \eqref{u} achieves practical prescribed-time convergence of the position error $\mathbf{x}_{e}$, while ensuring it always remains within the safe tube defined by \eqref{tube}. 
\end{theorem}

\textit{Proof}. We analyze the closed-loop behavior over two time intervals $[0,T_f^\ast)$ and $[T_f^\ast,\infty)$, separately.

1) We first consider the time interval $t\in[0,T_f^\ast)$. Substituting \eqref{u} into \eqref{dx_e} yields
\begin{equation}
\label{dx_e1}
\dot{\mathbf{x}}_e = -k_1\alpha_f(t)\mathbf{x}_e-k_2\mathbf{z}+\mathbf{R}(\theta)\mathbf{u}_d.
\end{equation}
To proceed, let us define
\begin{subequations}
\begin{align}
\bar{\mathbf{x}}_e(s)&:=\mathbf{x}_e(\eta_f(s)), \label{x_e_trans}\\
\bar{\xi}(s)&:=\xi(\eta_f(s))=\|\bar{\mathbf{x}}_{e}(s)\|^2/\rho^2, \\
\bar{\mathbf{z}}(s)&:=\mathbf{z}(\eta_f(s))=\bar{\mathbf{x}}_e(s)/(\rho^2(1-\bar{\xi}(s))).
\end{align}
\end{subequations}
Further define $\bar{\mathbf{d}}(s):=\mathbf{R}(\theta(\eta_f(s)))\mathbf{u}_d(\eta_f(s))$ for notational concision. Then, the closed-loop system \eqref{dx_e1} can be rewritten in the stretched infinite time interval $s\in[0,\infty)$ as
\begin{align}
\label{dbar_x_e}
\bar{\mathbf{x}}_e^\prime(s)&=\dfrac{d\bar{\mathbf{x}}_e(s)}{ds}=\dfrac{d\mathbf{x}_e(\eta_f(s))}{d\eta_f(s)}\cdot\dfrac{d\eta_f(s)}{ds} \nonumber \\
              &=-k_1\bar{\mathbf{x}}_e(s)-\dfrac{1}{\alpha_f(\eta_f(s))}(k_2\bar{\mathbf{z}}(s)-\bar{\mathbf{d}}(s)),
\end{align}
with $\bar{\mathbf{x}}_e(0)=\mathbf{x}_e(0)$ and $\bar{\mathbf{x}}_e^\prime(0)=\dot{\mathbf{x}}_e(0)$.

Consider the following logarithmic barrier function
\begin{equation}
\label{L}
L(\bar{\xi})=\dfrac{1}{2}\ln\dfrac{1}{1-\bar{\xi}}. 
\end{equation}
Taking the derivative of $L$ w.r.t. $s$ along \eqref{dbar_x_e} yields
\begin{equation}
\label{dL}
L^\prime(\bar{\xi}(s))=\bar{\mathbf{z}}^{\top}(s)\left(-k_1\bar{\mathbf{x}}_e(s)-\dfrac{k_2\bar{\mathbf{z}}(s)-\bar{\mathbf{d}}(s)}{\alpha_f(\eta_f(s))}\right).
\end{equation}
From \cite[Lemma 3]{shao2023fault}, it follows that $-k_1\bar{\mathbf{z}}^{\top}\bar{\mathbf{x}}_{e}=-k_1\bar{\xi}/(1-\bar{\xi})\leq-k_1L$ holds for all $0\leq\bar{\xi}<1$. In addition, recalling Assumption \ref{A1} and the fact that $\|\mathbf{R}(\theta)\|=1$, we have $\bar{\mathbf{z}}^\top\bar{\mathbf{d}}\leq k_2\|\bar{\mathbf{z}}\|^2+d/(4k_2)$. As a result, \eqref{dL} reduces to 
\begin{align}
\label{dL1}
L^\prime(\bar{\xi}(s))&\leq -k_1L(\bar{\xi}(s))+\dfrac{d}{4k_2}\dfrac{T_f-\eta_f(s)}{T_f} \nonumber \\
&=-k_1L(\bar{\xi}(s))+\dfrac{d}{4k_2}e^{-\frac{s}{T_f}},
\end{align}
on the set $0\leq\bar{\xi}<1$. Integrating both sides of \eqref{dL1} leads to
\begin{align}
\label{L1}
L(\bar{\xi}(s)) &\leq L(\bar{\xi}(0))e^{-k_1s}+De^{-\frac{s}{T_f}},
\end{align}
where $D:=T_fd/(4k_2(k_1T_f-1))>0$ is a constant. From \eqref{L1}, it is clear that $L(\bar{\xi})$ is bounded, indicating that $0\leq \bar{\xi}(s) <1$ for all $s\geq0$. As a result, $\|\bar{\mathbf{x}}_e(s)\|<\rho$ holds for all $s\geq0$. Since $t\to T_f$ as $s\to\infty$, it follows from \eqref{x_e_trans} that $\mathbf{x}_e(t)$ keeps within the safe tube \eqref{tube} for all $t\in[0,T_f^\ast)$. 

In the following, we show that $\mathbf{x}_e(t)$ converges to a small neighborhood of origin at $t=T_f^\ast$. It follows from \eqref{L1} that $\lim_{s\to s^\ast}L(\bar{\xi}(s)) \leq L^\ast:=L(\bar{\xi}(0))e^{-k_1s^\ast}+De^{-s^\ast/T_f}$. As per the definition of $\eta_f(s)$, one can easily deduce that $s \to s^\ast:=-T_f\ln(\varsigma_f/T_f)$ as $t\to T_f^\ast$. Since $0<\varsigma_f\ll T_f$, $s^\ast$ is a  very large value, which renders $L^\ast$ very small. By \eqref{xi} and \eqref{L}, we get $\lim_{s\to s^\ast}\|\bar{\mathbf{x}}_{e}(s)\|\leq \rho\sqrt{1-e^{-2L^\ast}}$, which implies that $\bar{\mathbf{x}}_{e}$ converges to a small neighborhood of origin as $s\to s^\ast$. Since $t\to T_f^\ast$ as $s \to s^\ast$ and $\bar{\mathbf{x}}_e(s)=\mathbf{x}_e(\eta_f(s))$ (see \eqref{x_e_trans}), it can be claimed that $\lim_{t\to T_f^\ast}\|\mathbf{x}_{e}(t)\|\leq \rho\sqrt{1-e^{-2L^\ast}}$.


2) Consider the time interval $t\in[T_f^\ast,\infty)$, in which $\alpha_f^\ast(t)=\alpha_f(T_f^\ast)\equiv T_f/\varsigma_f$ and the closed-loop system becomes
\begin{equation}
\label{dx_e2}
\dot{\mathbf{x}}_e = -k_1\alpha_f(T_f^\ast)\mathbf{x}_e-k_2\mathbf{z}+\mathbf{R}(\theta)\mathbf{u}_d.
\end{equation}
We likewise consider the barrier function $L$ in \eqref{L}. Taking its derivative w.r.t. $t$ along \eqref{dx_e2} gets
\begin{align}
\label{dL4}
\dot{L}(\xi(t))\leq-\bar{k}L(\xi(t))+\bar{d},
\end{align}
where $\bar{k}:=k_1T_f/\varsigma_f$ and $\bar{d}:=d/(4k_2)$ are defined for notational concision. Solving \eqref{dL4}, we get
\begin{align}
\label{L2}
L(\xi(t))\leq L(\xi(T_f^\ast))e^{-\bar{k}(t-T_f^\ast)}+\bar{d}/\bar{k},~\forall t\geq T_f^\ast,
\end{align}
indicating that $L(\xi(t))$ is bounded on $t\in[T_f^\ast,\infty)$. Thus, $\mathbf{x}_e(t)$ satisfies the tube constraint \eqref{tube} for all $t\geq T_f^\ast$. In addition, $L$ remains within the set $L(\xi(t))\leq L^\ast+\bar{d}/\bar{k}$ for all $t\geq T_f^\ast$. As a result, the tracking error $\mathbf{x}_e$ remains within the residual set $\|\mathbf{x}_e(t)\|\leq\rho\sqrt{1-e^{-2(L^\ast+\bar{d}/\bar{k})}}$ for all $t\geq T_f^\ast$. 

Based on the above analyses, we conclude that the tracking error $\mathbf{x}_e$ converges to a small residual set within the prescribed time $T_f^\ast$, while remaining within the safe tube defined by \eqref{tube} for all times. This completes the proof. $\hfill \blacksquare$ 	

\begin{remark}
	Under the InPTC framework, if the radius of the tube is set less than the safety margin $\epsilon$, whilst $T_f$ is set no larger than $T$, then the actual trajectory of $\mathbf{x}$ will converge to a small neighborhood of the goal position $\mathbf{x}^\ast$ (if no local minima occurs) within the prescribed time $T$, while avoiding collision with all obstacles $\mathcal{O}_i$, $i\in\mathbb{I}$ along the way. Therefore, the proposed InPTC can achieve prescribed-time collision-free navigation of WMRs. Furthermore, inspecting \eqref{L1} and \eqref{L2} reveals that choosing a smaller $\varsigma_f$ together with larger values for $k_1$ and $k_2$ decreases the size of residual set. However, this comes at the price of higher control gains (see \eqref{u}), particularly when selecting a very small $\varsigma_f$. This, in turn, may excite the unmodeled high frequency dynamics, leading to instability of the closed system. Thus, $k_1$, $k_2$, and $\varsigma_f$ should be judiciously chosen to strike a balance between tracking accuracy and system stability. 
\end{remark}



\section{Simulation Results} \label{secIV}

In this section, we demonstrate the proposed InPTC scheme in a $6.4\,\text{m}\times3.4\,\text{m}$ rectangular workspace $\mathcal{W}^{\ast}$ cluttered with 8 circular obstacles, denoted as $\mathcal{O}_{i}^{\ast}$, whose centers and radii are listed in Table \ref{tab1}. The distance between the virtual and actual control points is $\ell=0.05\,\rm m$, the radius of the circumscribed circle around $\bm{P}$ is $r=0.2\,\rm m$. The safety margin and influence region of obstacles are set to $\epsilon=0.1\,\rm m$ and $\epsilon^{\ast}=0.2\,\rm m$, respectively. The desired goal is $\mathbf{x}^{\ast}=[2.5,1]^{\top}\,\rm m$, and the task completion time is $200\,\rm s$. The disturbance $\mathbf{u}_d$ is modeled as $\mathbf{u}_d=0.01[\sin(0.2t)+1,\cos(0.3t)-2]^\top$. The simulation duration is $1000\,\rm s$, and the sample step is $0.05\,\rm s$.


\begin{table}
	\centering
	\caption{Geometrical details of the obstacles. \label{tab1}}
	\setlength\tabcolsep{4pt} 
	\renewcommand{\arraystretch}{1.3}  
	\begin{tabular}{|c|c|c|c|c|c|}
		\hline
		\multirow{2}{*}{Index} &
		\multicolumn{2}{c|}{Configuration} & \multirow{2}{*}{Index} &
		\multicolumn{2}{c|}{Configuration} \\
		\cline{2-3} \cline{5-6} 
		& Center (m) & Radius (m) &&  Center (m) & Radius (m) \\
		\hline
		$\mathcal{O}_{1}^{\ast}$ & $[-2,-0.55]^{\top}$ & $0.10$ & $\mathcal{O}_{5}^{\ast}$ & $[0.4,0.55]^{\top}$ & $0.25$\\
		
		$\mathcal{O}_{2}^{\ast}$ & $[-0.9,0.85]^{\top}$ & $0.10$ & $\mathcal{O}_{6}^{\ast}$ & $[0.7,-0.6]^{\top}$ & $0.10$\\
		
		$\mathcal{O}_{3}^{\ast}$ & $[-0.7,-0.5]^{\top}$ & $0.35$ & $\mathcal{O}_{7}^{\ast}$ & $[2,-0.6]^{\top}$ & $0.25$\\
		
		$\mathcal{O}_{4}^{\ast}$ & $[-2.1,0.6]^{\top}$ & $0.15$ & $\mathcal{O}_{8}^{\ast}$ & $[1.8,0.7 ]^{\top}$ & $0.15$\\
		\hline
	\end{tabular}
\end{table}

We begin by illustrating the effectiveness of the proposed prescribed-time planner \eqref{h_ast1} (denoted as ``PTP''). For comparison purposes, the APF-based planner presented in \cite{wang2024hybrid} (denoted as ``APF'') and the CBF-based planner presented in \cite{singletary2021comparative} (denoted as ``CBF'') are also simulated. The attractive and repulsive potentials of the APF planner are designed as $U_{att}(\mathbf{x}_{d})=\frac{k_a}{2}\|\mathbf{x}_{d}-\mathbf{x}^{\ast}\|^2$, $U_{rep}(\mathbf{x}_{d})=k_{r}\sum_{i=1}^{8}\Upsilon(\text{d}_{\mathcal{O}_{i}}(\mathbf{x}_d))$,
where $k_{a}$ and $k_{r}$ are positive weights, and $\Upsilon$ is a smooth repulsive function defined as $\Upsilon(z)=(\epsilon^\ast-z)^2\text{ln}(z-\epsilon)/(z-\epsilon)$ if $z\in(\epsilon,\epsilon^\ast]$, and $\Upsilon(z)=0$, if $z>\epsilon^\ast$. Then, the vector field generated by the APF planner is
\begin{equation}
\label{apf}
\bm{\tau}_{d}=-\nabla U_{att}(\mathbf{x}_{d}) -\nabla U_{rep}(\mathbf{x}_{d}).
\end{equation}
We consider a CBF $f(\mathbf{x}_d):=\min_{i\in\{0\}\cup\mathbb{I}}f_i(\mathbf{x}_d)$ together with the safe set $\mathcal{F}:=\{\mathbf{x}_{d}\in\mathbb{R}^2\mid f(\mathbf{x}_d)\geq0\}$, where $f_0(\mathbf{x}_d):=1-(x_d/2.9)^{20}-(y_d/1.4)^{20}$ (for workspace boundaries), and $f_i(\mathbf{x}_d):=\|\mathbf{x}_d-\mathbf{c}_i\|^2-(r+r_i+\epsilon)^2$, $i\in\mathbb{I}$. The CBF planner is obtained by quadratic programming
\begin{align*}
\mathop{\rm argmin} \limits_{\bm{\tau}_d \in \mathbb{R}^2}&~\|\bm{\tau}_d-\bm{\tau}_{d,\rm des}(\mathbf{x}_d)\|^2  \tag{CBF-QP}\\
\text{s.t.} &~ \nabla f(\mathbf{x}_d)^\top\bm{\tau}_d \geq - \gamma f(\mathbf{x}_d),
\end{align*}
where $\alpha>0$ is a user-defined constant. The CBF-QP has an explicit solution of the form
\begin{equation}
\label{cbf}
\bm{\tau}_d = \bm{\tau}_{d,\rm des}(\mathbf{x}_d) + \bm{\tau}_{d,\rm safe}(\mathbf{x}_d),
\end{equation}
where $\bm{\tau}_{d,\rm des}(\mathbf{x}_d):=\bm{\kappa}_0(\mathbf{x}_d)$, and $\bm{\tau}_{d,\rm safe}$ is determined by 
\begin{equation*}
\bm{\tau}_{d,\rm safe}(\mathbf{x}_d):= \left\lbrace
\begin{aligned}
&-\dfrac{\nabla f(\mathbf{x}_d)}{\nabla f(\mathbf{x}_d)^\top\nabla f(\mathbf{x}_d)}\Psi(\mathbf{x}_d),&~&\text{if}~\Psi<0, \\
&0,&~&\text{if}~\Psi\geq0,
\end{aligned}\right.
\end{equation*}
with $\Psi(\mathbf{x}_d):=\nabla f(\mathbf{x}_d)^\top\bm{\tau}_{d,\rm des}(\mathbf{x}_d)+\gamma f(\mathbf{x}_d)$. As dictated by \cite{singletary2021comparative}, the CBF planner \eqref{cbf} ensures that the set $\mathcal{F}$ is forward invariant. The APF planner parameters are chosen as $k_{a}=k_0$ and $k_{r}=0.1$, while the CBF planner parameter is chosen as $\gamma=0.1$. It is important to note that these three planners have the same motion-to-goal control law. Their design parameters are listed in Table \ref{tab2}.

\begin{table}
	\caption{Planner and controller parameters.\label{tab2}}
	\centering
	\begin{threeparttable}
		\begin{tabular*}{250pt}{@{\extracolsep\fill}lcc@{\extracolsep\fill}}%
			\toprule
			Method      &     Parameter     \\
			\midrule
			PTP in \eqref{h_ast1}        &     $k_0=0.01$, $T=200$, $\varsigma=0.5$                 \\
			TFC in \eqref{u}  &     $\rho=0.06$, $k_1=0.8$, $k_2=0.001$, $T_f=200$, $\varsigma_f=3$                     \\
			APF in \cite{wang2024hybrid}           &     $k_{a}=k_0=0.01$, $k_{r}=0.1$         \\
			CBF in \cite{singletary2021comparative}         &     $\gamma=0.1$             \\ 
			\bottomrule
		\end{tabular*}
	\end{threeparttable}
\end{table}

\begin{figure}[htb!]
	\centering
	\includegraphics[width=8.8cm]{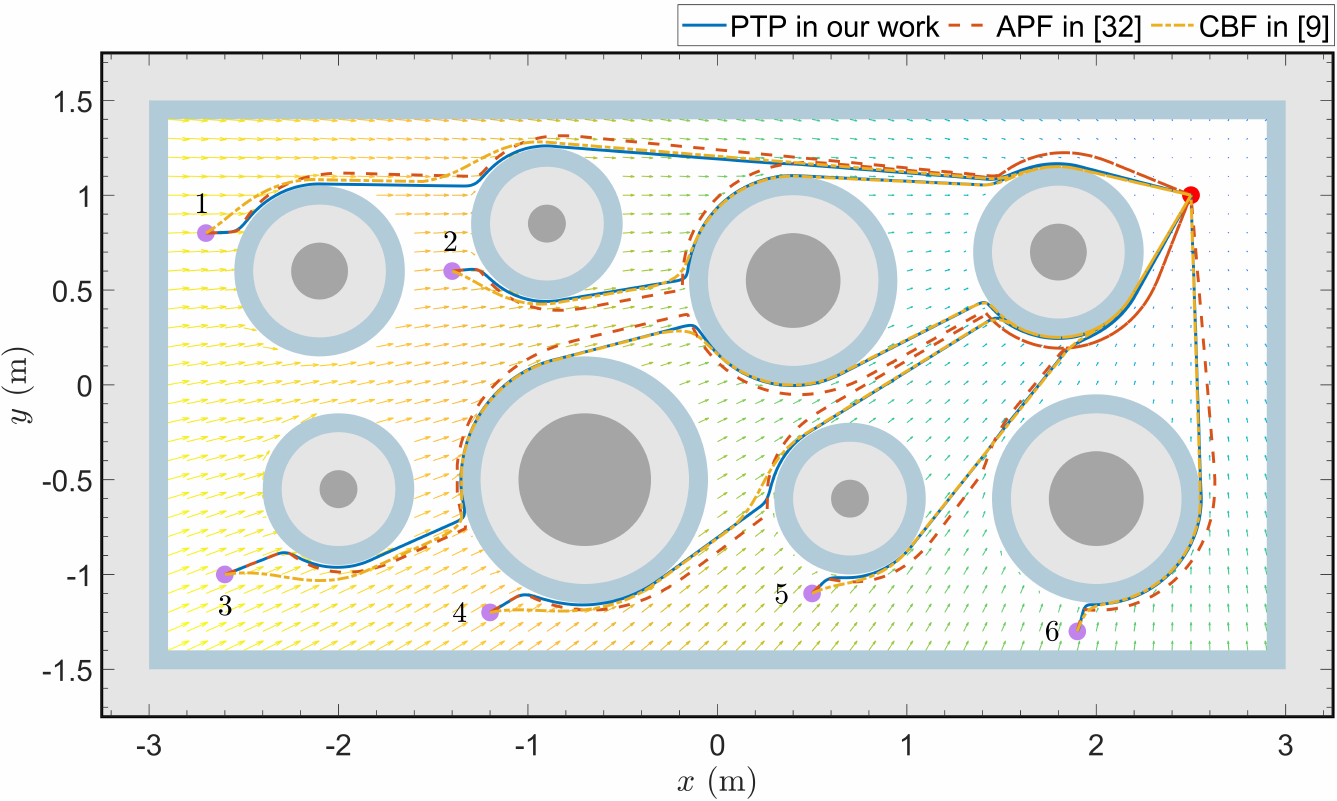}
	\caption{Planned trajectories of the three path planners.}
	\label{2D_trajectory}
\end{figure}

\begin{figure}[!htbp]
	\centering
	\subfigure[PTP in our work]{
		\includegraphics[width=8.8cm]{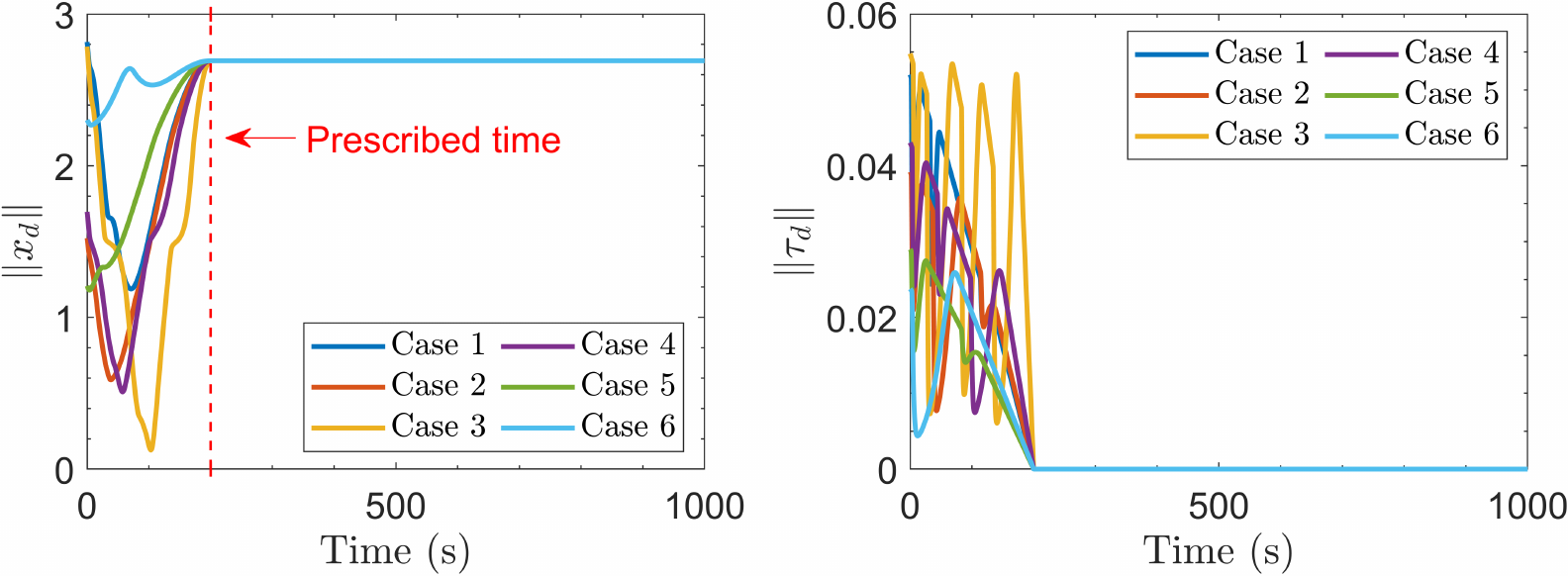}
		\label{control_response.a}}
	\subfigure[APF in \cite{wang2024hybrid}]{
		\includegraphics[width=8.8cm]{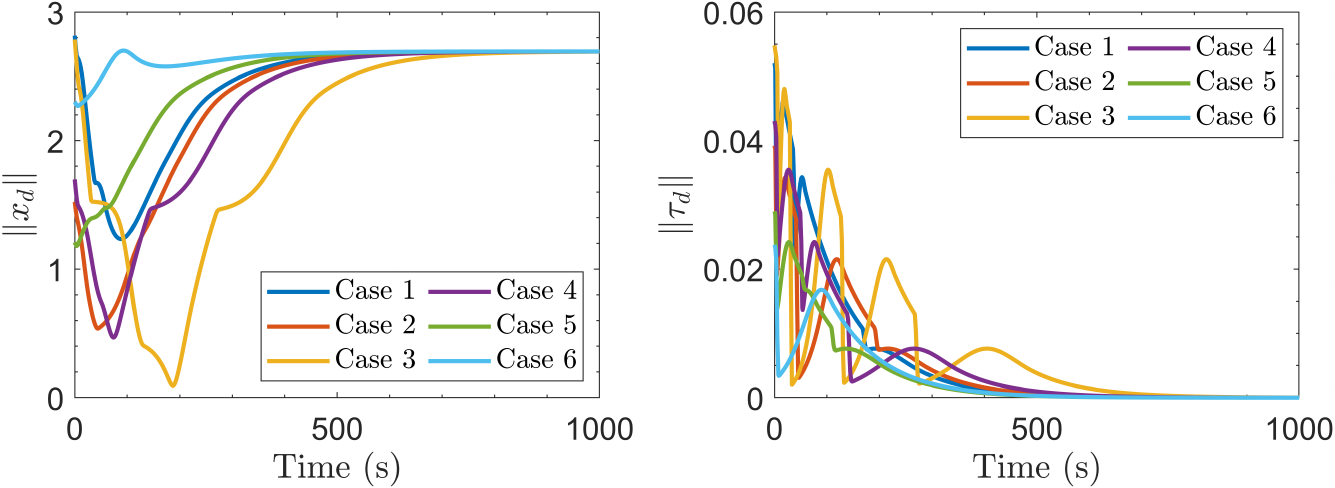}
		\label{control_response.b}}
	\subfigure[CBF in \cite{singletary2021comparative}]{
		\includegraphics[width=8.8cm]{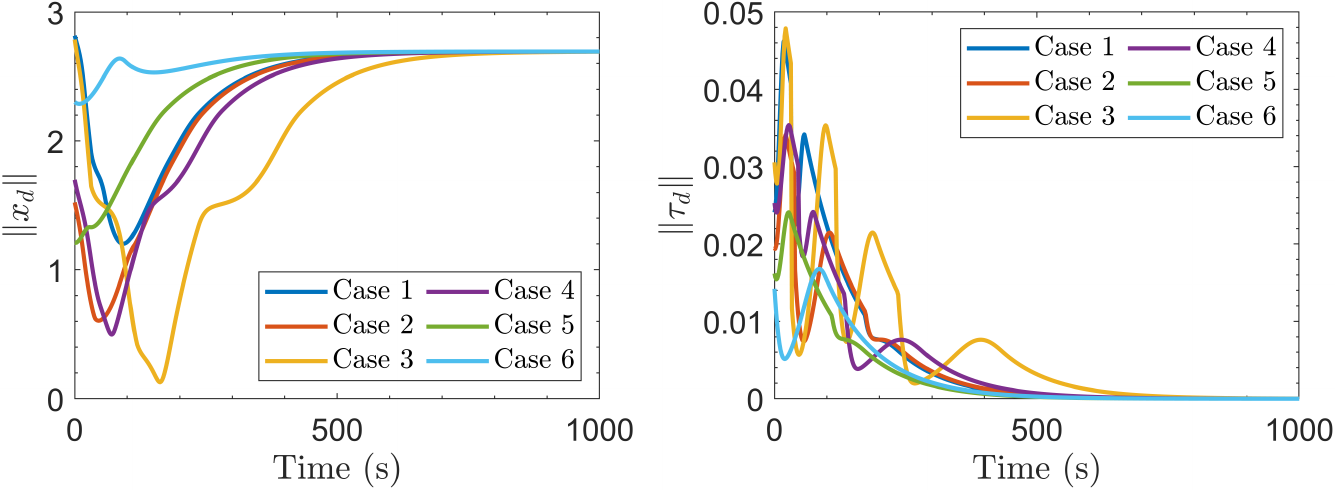}
		\label{control_response.c}}
	\caption{Control responses of different path planners.}
	\label{control_response}
\end{figure}

The planned trajectories starting at a set of initial positions (purple points) are plotted in Fig. \ref{2D_trajectory}, where the dense arrows denote the magnitude and direction of the resulting vector field from the prescribed-time planner \eqref{h_ast1} at various points. These vector arrows are directly obtained as per \eqref{h_ast1}. As shown in Fig. \ref{2D_trajectory}, all three planners successfully guide the WMR from different initial positions to the goal (red point), while avoiding all obstacle regions. The comparison results of $\|\mathbf{x}_{d}\|$ and $\|\bm{\tau}_{d}\|$ are depicted in Fig. \ref{control_response}, from which we find that both the APF and CBF converge slower than the proposed PTP; moreover, their convergence times become longer as the path is longer. However, the PTP always converges at $T=200\,\rm s$, regardless of initial positions. We further provide quantitative comparison results in Table \ref{tab3}, where the 3rd initial position is considered as a case study. Although all the three planners offer analytical solutions and have an almost identical time consumption (average time of 100 runs), the convergence time differs greatly. The PTP converges at the prescribed time $T=200\,\rm s$, while the APF and CBF take significantly longer, converging in $950\,\rm s$. The path length metric also differs, with CBF covering a shorter path compared to the other two methods. The maximum and standard deviation (std) of the velocity norm provide additional insights, with PTP exhibiting higher values than APF and CBF in both measures, due to its faster convergence rate.
 
%
%

\begin{table*}
	\caption{Planning performance comparison.\label{tab3}}
	\centering
	\begin{threeparttable}
		\begin{tabular*}{500pt}{@{\extracolsep\fill}lccccccc@{\extracolsep\fill}}%
			\toprule
			Method      &     Time consumption    &    Convergence time  & Path length & Maximum of $\|\bm{\tau}_{d}\|$ & std of $\|\bm{\tau}_{d}\|$ & Obstacle avoidance & Prescribed time  \\
			\midrule
			PTP in \eqref{h_ast1}        &     0.3136\,s     &    200\,s &    6.4411\,m &    0.0548 &   0.0148 & \Checkmark & \Checkmark  \\
			APF in \cite{wang2024hybrid}   &   0.3196\,s    &    950\,s  &     6.7394\,m       &    0.0548 &    0.0096 & \Checkmark & \XSolidBrush   \\
			CBF in \cite{singletary2021comparative}         &     0.3474\,s    &    950\,s  &     6.3976\,m    &    0.0479 &    0.0102 & \Checkmark & \XSolidBrush  \\ 
			\bottomrule
		\end{tabular*}
	\end{threeparttable}
\end{table*}

\begin{figure}[htb!]
	\centering
	\includegraphics[width=8.5cm]{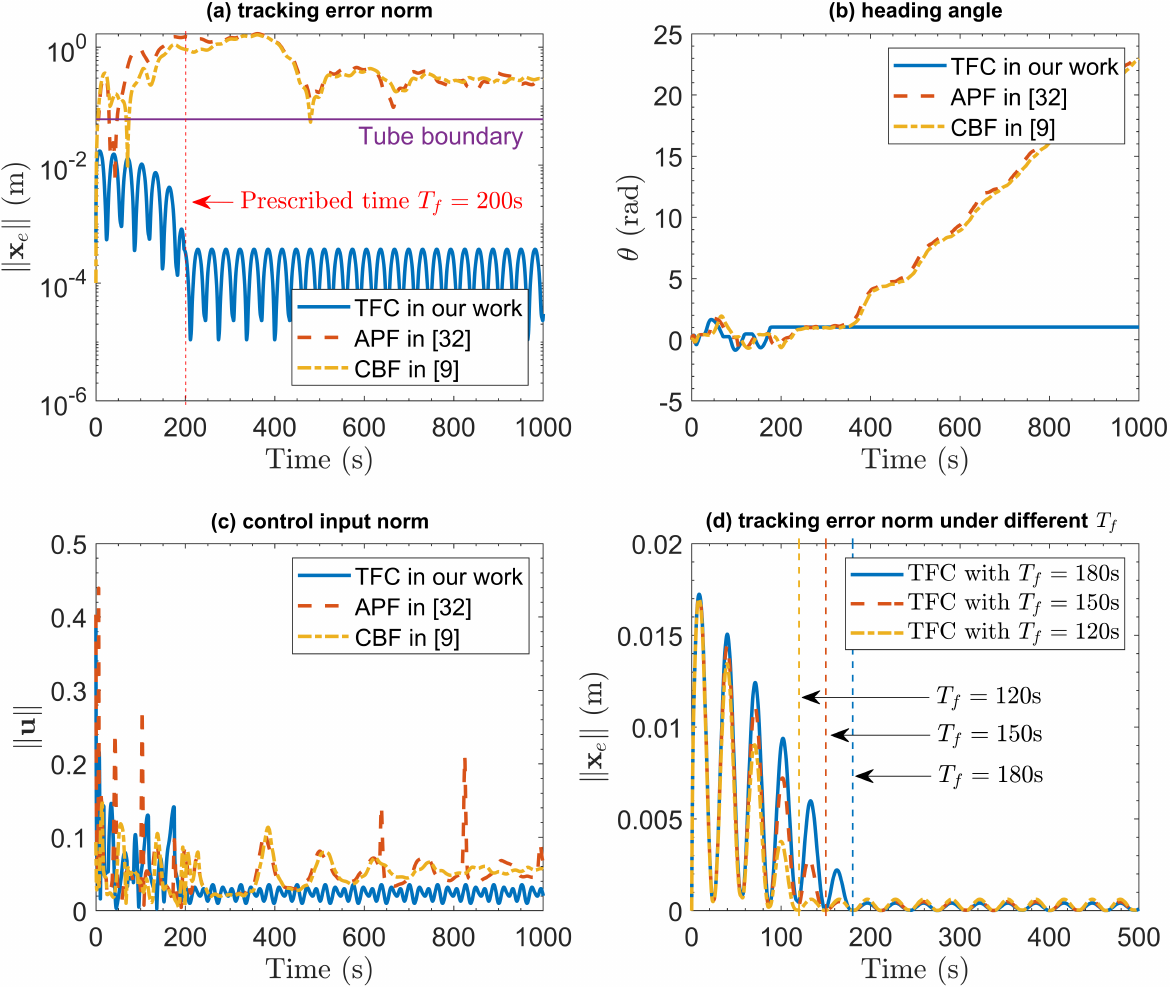}
	\caption{Comparisons of the closed-loop tracking responses.}
	\label{tracking_comparison}
\end{figure}

\begin{figure}[htb!]
	\centering
	\includegraphics[width=8.5cm]{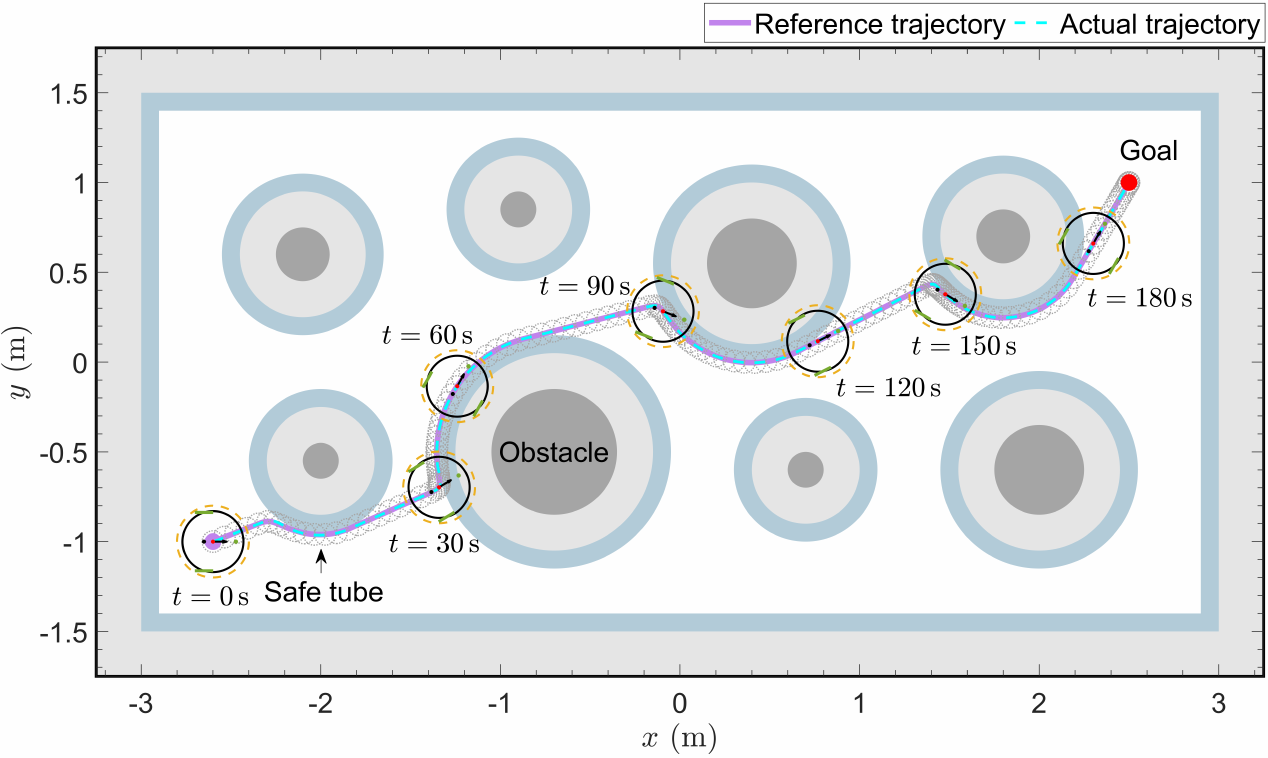}
	\caption{Position tracking trajectory with WMR snapshots.}
	\label{tracking_trajectory}
\end{figure}

In the following, we show the effectiveness and performance of the proposed tube-following controller \eqref{u} (denoted as ``TFC'') with parameters given in Table \ref{tab2}. As a case study, the planned trajectory starting from the 3rd initial position is selected as the reference trajectory. To verify the robustness of the APF and CBF planners, we execute them in the presence of disturbances and define the difference between the reference and actual trajectories as the position tracking error $\mathbf{x}_{e}$. The closed-loop tracking responses of three controllers are shown in Fig. \ref{tracking_comparison}. In Fig. \ref{tracking_comparison}(a), the tracking error $\mathbf{x}_{e}$ under the proposed TFC is shown to converge to a residual set of $3.74\times10^{-4}\,\rm m$ after the prescribed time $T_f=200\,\rm s$, while complying with the tube constraint defined by \eqref{tube}. In contrast, the APF and CBF controllers exhibit larger stead-state errors of $0.43\,\rm m$ and $0.32\,\rm m$, respectively, whilst violate the tube constraint, leading to an increased risk of collision. The WMR's heading angle is provided in Fig. \ref{tracking_comparison}(b), from which it is clear that $\theta$ under the TFC is bounded and converges to a nearly constant value, while under the APF and CBF, $\theta$ continuously increases. This implies that both the APF and CBF controllers are susceptible to disturbances. The control norm $\|\mathbf{u}\|$ is shown in Fig. \ref{tracking_comparison}(c). To further show the prescribed-time convergence property of the proposed TFC, the tracking error norm under different prescribed time $T_f$ is given in Fig. \ref{tracking_comparison}(d). As illustrated, the proposed controller can precisely and flexibly set the convergence time in advance. The position tracking trajectory of the proposed TFC is shown in Fig. \ref{tracking_trajectory}. Intuitively, $\mathbf{x}$ accurately tracks the reference trajectory, while remaining within the safe tube. This implies that, under the proposed InPTC, the WMR reaches the goal within the prescribed time $T=200\,\rm s$, while avoiding collision with any obstacles (dark gray balls). 


\section{Experiments} \label{secV}

\begin{figure}[t]
	\centering
	\includegraphics[width=8.8cm]{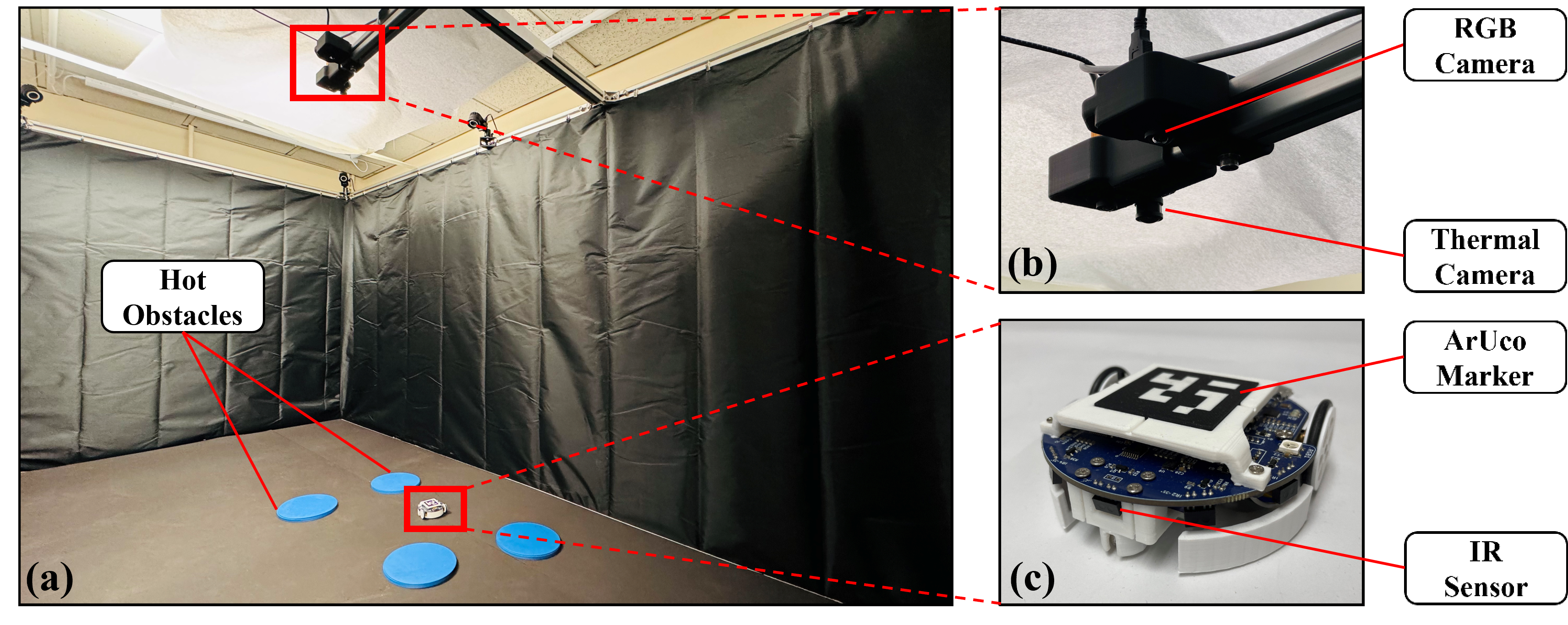}
	\caption{The experiment setup: (a) platform; (b) cameras; (c) Mona robot.}
	\label{fig10}
\end{figure}

The experimental platform is shown in Fig. \ref{fig10}, which consists of a Mona robot \cite{arvin2019mona} with nonholonomic dynamics, several hot obstacles containing iron powders that can heat up themselves, a RGB camera to observe the pose and location of the WMR, a thermal camera to obtain the size and location of the hot obstacles, and a control PC to collect feedback data and send motion commands. The Wi-Fi communication (at a rate of $10\,\rm Hz$) between the control PC and robot is built by ROS. All experiments are performed in a $\rm 2.78\,m \times 1.4\,m$ arena. Since the Mona robot has a very limited onboard sensing capability, the distance between the robot and obstacle boundary is obtained in real-time from the RGB and thermal cameras. The radius of the circle surrounding the robot is set to $r=0.06\,\text{m}$, the parameter for the change of coordinates is $\ell=-0.02\,\rm m$. The design parameters of the InPTC are listed in Table \ref{tab4}. In experiments, the disturbances could come from measurement noises, communication delay, imperfect wheel alignment, motor or actuator asymmetry, etc. In the following, we report the results of two representative experiments, where the extracted thermal image and the RGB image are merged to record the experimental process.

\begin{table}[t]
	\caption{Design parameters of the InPTC.\label{tab4}}
	\centering
	\begin{threeparttable}
		\begin{tabular*}{250pt}{@{\extracolsep\fill}lcc@{\extracolsep\fill}}%
			\toprule
			Method      &     Parameter     \\
			\midrule
			Planner in \eqref{h_ast1}        &     $k_0=0.01$, $T=250$, $\varsigma=0.5$                 \\
			Controller in \eqref{u}  &     $\rho=0.06$, $k_1=0.01$, $k_2=0.003$, $T_f=150$, $\varsigma_f=2$                     \\
			\bottomrule
		\end{tabular*}
	\end{threeparttable}
\end{table}

\begin{figure}[htb!]
	\centering
	\includegraphics[width=8.5cm]{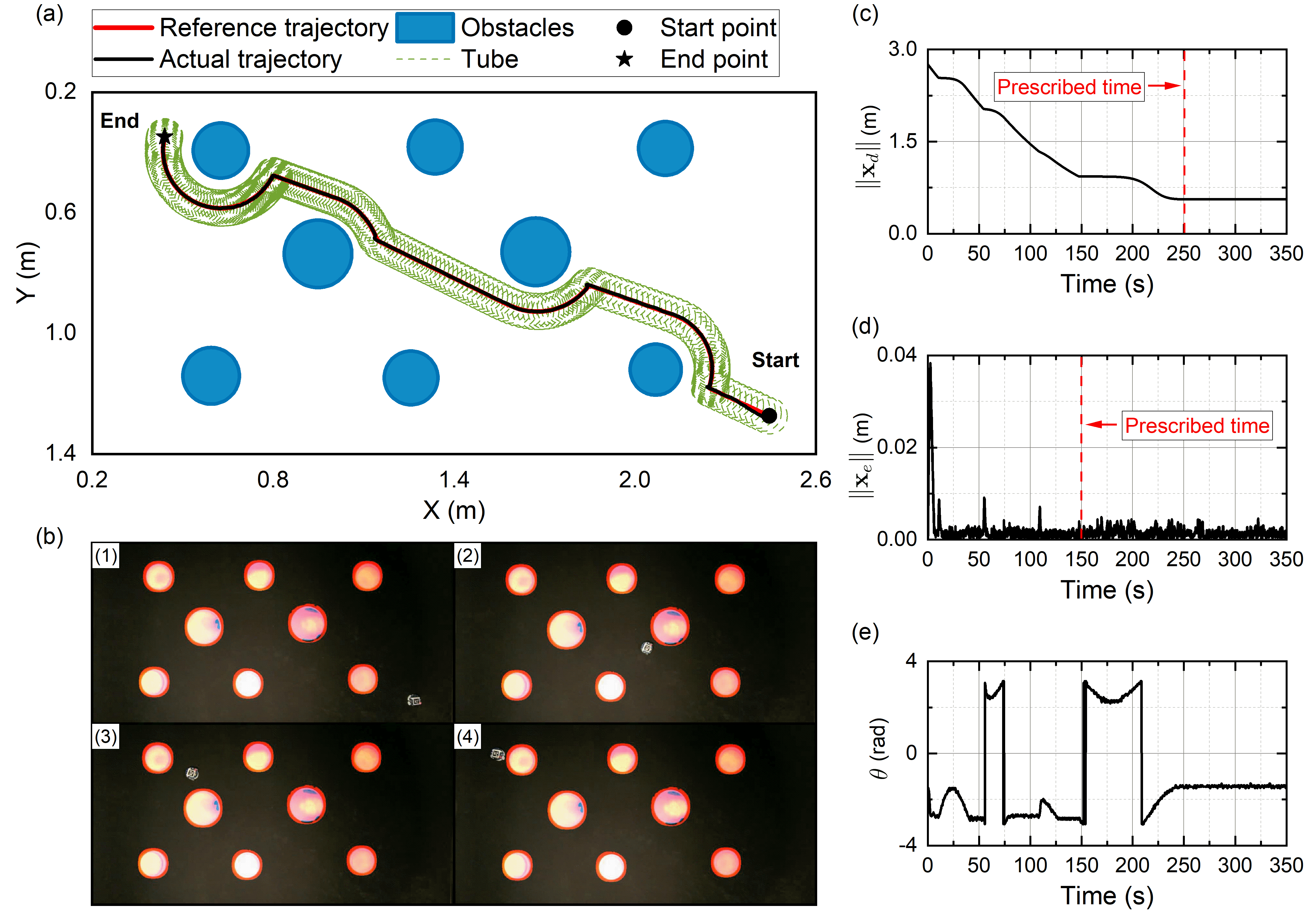}
	\caption{The experiment with circular obstacles: (a) WMR's motion trajectory; (b) screenshots of the experiment; (c) reference position norm; (d) position error norm; (e) WMR's heading angle.}
	\label{fig11}
\end{figure}

In the first experiment, eight circular obstacles are placed in the arena, and the safety margin and influence region of obstacles are set to $\epsilon=0.04\,\text{m}$ and $\epsilon^*=0.05\,\text{m}$, respectively. The WMR starts from the right side of the arena (coordinates: $[2.45, 1.27]^{\top}\text{m}$), and is tasked with reaching the goal located on the right side (coordinates: $[0.44, 0.35]^{\top}\text{m}$). The task is required to be completed within $250\,\rm s$. The results are shown in Fig. \ref{fig11}, from which it is clear that the proposed planner in \eqref{h_ast1} generates a smooth, collision-free reference trajectory that converges to the goal at the preassigned time $T=250\,\rm s$ (i.e., the red line in Fig. \ref{fig11}(a)). While the proposed controller in \eqref{u} achieves trajectory tracking along with the safe tube (see Fig. \ref{fig11}(c)), with a tracking accuracy higher than $4.9 \times 10^{-3}\,\rm m$ after the preassigned time $T_f=150\,\rm s$ (see Fig. \ref{fig11}(d)). As seen in Fig. \ref{fig11}(e), the robot's heading angle remains nearly constant after the task completion time $T=250\,\rm s$. The above results demonstrate that the proposed InPTC scheme enables the WMR to safely accomplish the navigation task within the required time of $250\,\rm s$.

\begin{figure}[htb!]
	\centering
	\includegraphics[width=8.5cm]{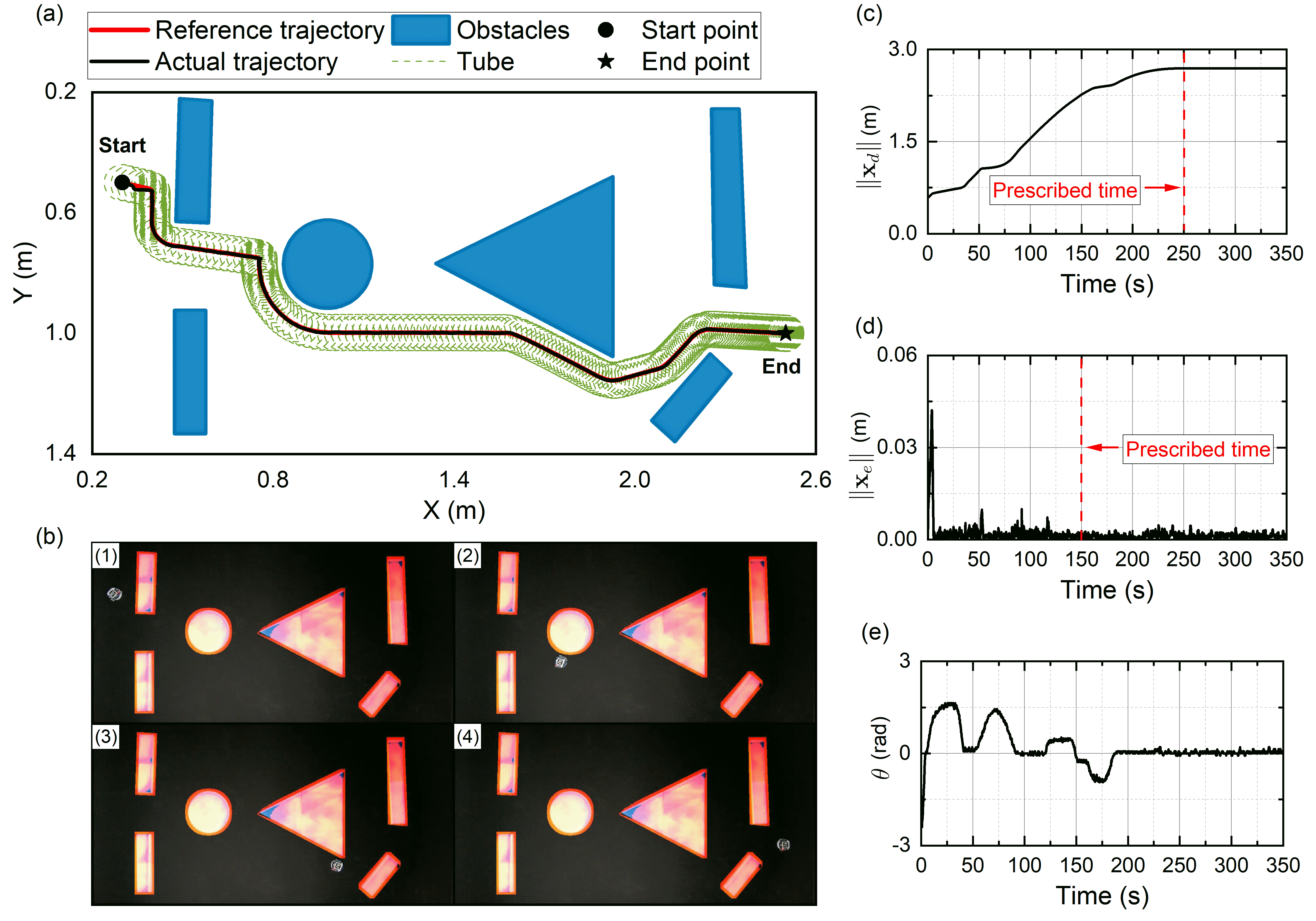}
	\caption{The experiment with circular and polygonal obstacles: (a) WMR's motion trajectory; (b) screenshots of the experiment; (c) reference position norm; (d) position error norm; (e) WMR's heading angle.}
	\label{fig12}
\end{figure}

In the second experiment, we consider a more complex environment with six obstacles of various shapes. The safety margin and influence region of obstacles are set to $\epsilon=0.08\,\text{m}$ and $\epsilon^*=0.1\,\text{m}$, respectively.
The WMR starts from the left side of the arena (coordinate: $[0.3, 0.5]^{\top}\text{m}$) and is required to reach a goal location at the right side (coordinate: $[2.5, 1.0]^{\top}\text{m}$). The results are depicted in Fig. \ref{fig12}, hich shows outcomes analogous to those observed in the initial experiment. In this case, the proposed InPTC still generates a smooth and safe reference trajectory with prescribed-time convergence (see Fig. \ref{fig12}(c)). It also achieves prescribed-time tube-following control with an accuracy of $4.7 \times 10^{-3}\,\rm m$ (see Fig. \ref{fig12}(d)), thus enabling the WMR to accomplish the navigation task within the prescribed time $T=250\,\rm s$ (see Fig. \ref{fig12}(a)).

Furthermore, we perform multiple trials with different initial and goal locations to further test the proposed InPTC. From these experiments, we consistently observe the effectiveness of InPTC in achieving tracking accuracy higher than $0.01\,\rm m$ and task completion time less than $250\,\rm s$. More details are provided in the supplementary video (\url{https://vimeo.com/895801720}). 


\section{Conclusion} \label{conclusion}

In this article, we propose a prescribed-time path planning algorithm for wheeled mobile robots (WMRs) to generate a collision-free reference trajectory that can converge to the goal location at a prescribed finite time. The derived planner is then paired with a tube-following controller that allows WMRs to achieve high-precision trajectory tracking within a prescribed time, while remaining within a predefined ``safe tube'' around the reference trajectory. The resulting integrated planning and tube-following control (InPTC) scheme achieves prescribed-time collision-free navigation of WMRs, despite the presence of disturbances. The salient features of the proposed InPTC are two-fold: 1) it enables WMRs to accomplish safe navigation within a finite time that can be flexibly preassigned according to the mission requirements; and 2) it enhances the navigation safety of WMRs in practical implementations. Simulation and experimental results show the effectiveness of InPTC. Future work will focus on extending the proposed method to achieve prescribed-time safe navigation of WMRs in environments with dynamic or intersecting obstacles.

\appendix[Proof of Theorem {\ref{Th2}}]
	
 From \eqref{h} and \eqref{Pi_x}, it follows that $\mathbf{h}(\mathbf{x})\in\mathbf{T}_{\mathcal{X}_{\epsilon}}(\mathbf{x})$ for all $\mathbf{x}\in\mathcal{X}_{\epsilon}$, which according to Theorem \ref{Th1} suggests that the free space $\mathcal{X}_{\epsilon}$ is forward invariant. 

\begin{figure}[t]
 	\centering
 	\includegraphics[width=4.5cm]{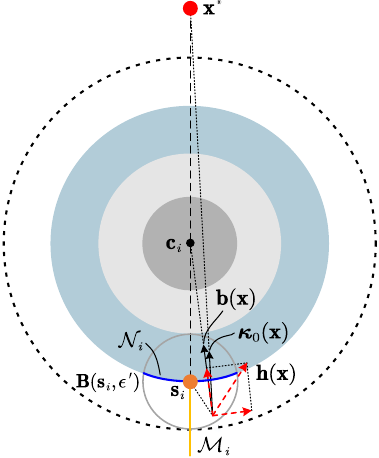}
 	\caption{Illustration of the undesired stationary point $\mathbf{s}_{i}$.}
 	\label{figA}
 \end{figure}

 As previously stated, $\mathbf{h}(\mathbf{x})$ is piecewise continuously differentiable, indicating that it is locally Lipschitz on its domain $\mathcal{X}$ \cite{chaney1990piecewise}. Moreover, the free space $\mathcal{X}{\epsilon}$, as a compact subset of $\mathcal{X}$, is forward invariant. Then, it follows from \cite[Theorem 3.3]{2002Nonlinear} that the closed-loop kinematics \eqref{dx} under $\bm{\tau}=\mathbf{h}(\mathbf{x})$ with $\mathbf{x}(0)\in\mathcal{X}_{\epsilon}$ has a unique solution that is defined for all $t\geq0$. We next show that the unique solution converges to a set of stationary points. To this end, we consider the following LFC:
 \begin{equation}
 \label{W}
 W(\mathbf{x})=\dfrac{1}{2}\|\mathbf{x}-\mathbf{x}^{\ast}\|^{2}.
 \end{equation} 
 For ease of analysis, \eqref{h} is rewritten as
 \begin{equation}
 \label{h_1}
 \mathbf{h}(\mathbf{x})=\underbrace{(c(\mathbf{x})\mathbf{I}_{2}+(1-c(\mathbf{x}))\mathbf{\Pi}(\mathbf{x}))}_{\mathbf{\Theta}(\mathbf{x})}\bm{\kappa}_{0}(\mathbf{x}),
 \end{equation}
 with
 \begin{equation}
 \label{c}
 c(\mathbf{x}):=\left\lbrace
 \begin{split}
 &1 ,&\text{if}~&\text{d}_{\mathcal{O}}(\mathbf{x})>\epsilon^{\ast}~\text{or}\\
 &&&\text{d}_{\mathcal{O}}(\mathbf{x})\leq\epsilon^{\ast}~\text{and}~\bm{\kappa}_{0}^{\top}(\mathbf{x})\mathbf{b}(\mathbf{x})\leq0, \\
 &0,&\text{if}~& \text{d}_{\mathcal{O}}(\mathbf{x})\leq\epsilon^{\ast}~\text{and}~ \bm{\kappa}_{0}^{\top}(\mathbf{x})\mathbf{b}(\mathbf{x})>0.
 \end{split}\right.
 \end{equation} 
 Taking the time derivative of $W(\mathbf{x})$ along \eqref{dx} and noting \eqref{kappa_0} and \eqref{h_1}, one gets
 \begin{equation}
 \label{dW}
 \dot{W}(\mathbf{x})=-k_0(\mathbf{x}-\mathbf{x}^{\ast})^{\top}\mathbf{\Theta}(\mathbf{x})(\mathbf{x}-\mathbf{x}^{\ast}).
 \end{equation} 
 Since $k_0>0$ and $\mathbf{\Pi}(\mathbf{x})$ is positive semidefinite, one has
 \begin{align}
 \label{dW_1}
 \dot{W}(\mathbf{x})\leq 0,~\forall \mathbf{x}\in\mathcal{X}_{\epsilon},
 \end{align}  
 indicating that $\mathbf{x}=\mathbf{x}^{\ast}$ is a stable equilibrium of \eqref{dx}. From LaSalle's invariance principle, it follows that the solution of \eqref{dx} asymptotically converges to the set $\{\mathbf{x}\in\mathcal{X}_{\epsilon}\mid\dot{W}(\mathbf{x})=0\}$, i.e., a set of stationary points. One can infer from \eqref{Pi_x} and \eqref{dW} that the stationary points correspond to either $\mathbf{x}=\mathbf{x}^{\ast}$ or the points satisfying $\text{d}_{\mathcal{O}}(\mathbf{x})=\epsilon \wedge\bm{\kappa}_{0}^{\top}(\mathbf{x})\mathbf{b}(\mathbf{x})/\|\bm{\kappa}_{0}(\mathbf{x})\|=1$. Therefore, for any $\mathbf{x}(0)\in\mathcal{X}_{\epsilon}$, the solution of \eqref{dx} converges to the set $\mathcal{C}=\{\mathbf{x}^{\ast}\}\cup_{i=1}^{n}\{\mathbf{s}_{i}\}$. The set contains $n+1$ stationary points, all of which are isolated due to Assumption \ref{A2}. Geometrically, the undesired stationary point $\mathbf{s}_{i}$ (on the boundary of $\mathcal{O}_{i}^{\epsilon}$) and $\mathbf{x}^{\ast}$ are collinear with $\mathbf{c}_{i}$, but located on the opposite sides of $\mathbf{c}_{i}$, as seen in Fig. \ref{figA}. As per \eqref{dW}, the analytical form of $\mathbf{s}_{i}$ is obtained as follows:
 \begin{equation}
 \label{s_i}
 \mathbf{s}_{i}=(1+\alpha_{i})\mathbf{c}_{i}-\alpha_{i}\mathbf{x}^{\ast},
 \end{equation}
 with $\alpha_{i}:=(r+r_{i}+\epsilon)\|\mathbf{x}^{\ast}-\mathbf{c}_{i}\|^{-1}$. 

 The isolated stationary points $\mathbf{s}_{i}$, $i\in\mathbb{I}$ may prevent us from achieving the objective (motion-to-goal). To check the stability of $\mathbf{s}_{i}$, we define a small neighborhood of $\mathbf{s}_{i}$, i.e., an open ball $\mathcal{B}(\mathbf{s}_{i},\epsilon^{\prime})$ with $0<\epsilon^{\prime}\leq\min\{\epsilon,\epsilon^{\ast}-\epsilon\}$, such that $\bm{\kappa}_{0}^{\top}(\mathbf{x})\mathbf{b}(\mathbf{x})>0$ holds for all $\mathbf{x}\in\mathcal{B}(\mathbf{s}_{i},\epsilon^{\prime})$. When restricted on $\mathcal{B}(\mathbf{s}_{i},\epsilon^{\prime})$, the feasible set of initial configurations is $\mathcal{F}_{i}:=\mathcal{X}_{\epsilon}\cap\mathcal{B}(\mathbf{s}_{i},\epsilon^{\prime})$, and $\mathbf{x}$ evolves according to 
 \begin{equation}
 \label{dx_2}
 \dot{\mathbf{x}}=-k_0(\mathbf{I}_{2}-\phi(\text{d}_{\mathcal{O}}(\mathbf{x}))\mathbf{b}(\mathbf{x})\mathbf{b}^{\top}(\mathbf{x}))(\mathbf{x}-\mathbf{x}^{\ast}).
 \end{equation}
 Further define two sets
 \begin{align}
 \mathcal{M}_{i}:=&~\left\lbrace \mathbf{x}\in\mathcal{X}_{\epsilon}\bigg|\text{d}_{\mathcal{O}_{i}}(\mathbf{x})\geq\epsilon,~\dfrac{\bm{\kappa}_{0}^{\top}(\mathbf{x})\mathbf{b}(\mathbf{x})}{\|\bm{\kappa}_{0}(\mathbf{x})\|}=1\right\rbrace, \label{M_i}\\
 \mathcal{N}_{i}:=&~\partial\mathcal{X}_{\epsilon}\cap\mathcal{B}(\mathbf{s}_{i},\epsilon^{\prime}). \label{N_i}
 \end{align}
 Geometrically speaking, all elements of $\mathcal{M}_{i}$ form a radial line extending outward from the stationary point $\mathbf{s}_{i}$ along $\mathbf{s}_{i}-\mathbf{c}_{i}$ (see the yellow line in Fig. \ref{figA}), while all elements of $\mathcal{N}_{i}$ form a curve, which is a portion of the boundary $\partial\mathcal{O}_{i}^{\epsilon}$ (see the blue curve in Fig. \ref{figA}). In the following, three cases are considered: 1) $\mathbf{x}(0)\in \mathcal{B}(\mathbf{s}_{i},\epsilon^{\prime})\cap\mathcal{M}_{i}$; 2) $\mathbf{x}(0)\in\mathcal{N}_{i}\setminus\{\mathbf{s}_i\}$; and 3) $\mathbf{x}(0)\in\mathcal{F}_{i}\setminus(\mathcal{M}_{i}\cup\mathcal{N}_{i})$. 

 Case 1): $\mathbf{x}(0)\in \mathcal{B}(\mathbf{s}_{i},\epsilon^{\prime})\cap\mathcal{M}_{i}$. Note that for all $\mathbf{x}\in\mathcal{M}_{i}$, $\bm{\kappa}_{0}^{\top}(\mathbf{x})\mathbf{b}(\mathbf{x})/\|\bm{\kappa}_{0}(\mathbf{x})\|=1$ holds, whereby \eqref{dx_2} reduces to
 \begin{equation}
 \label{dx_3}
 \dot{\mathbf{x}}=-k_0(1-\phi(\text{d}_{\mathcal{O}}(\mathbf{x})))(\mathbf{x}-\mathbf{x}^{\ast}).
 \end{equation}
 From \eqref{bump} and \eqref{M_i}, it is evident that $0\leq\phi(\text{d}_{\mathcal{O}}(\mathbf{x}))\leq1$ on the set $\mathcal{M}_{i}$ and $\phi(\text{d}_{\mathcal{O}}(\mathbf{x}))=1$ only when $\mathbf{x}=\mathbf{s}_{i}$. Hence, for all $\mathbf{x}\in\mathcal{M}_{i}\setminus\{\mathbf{s}_{i}\}$, the velocity vector $\mathbf{h}(\mathbf{x})$ (i.e., the right-hand side of \eqref{dx_3}) points directly towards $\mathbf{s}_{i}$ and becomes zero when $\mathbf{x}=\mathbf{s}_{i}$, indicating that the set $\mathcal{M}_{i}$ is forward invariant. Consider the following LFC:
 \begin{equation}
 \label{W_i}
 W_{i}(\mathbf{x})=\dfrac{1}{2}\|\mathbf{x}-\mathbf{s}_{i}\|^{2}.
 \end{equation}
 The time derivative of $W_{i}(\mathbf{x})$ along \eqref{dx_3} is given by 
 \begin{equation}
 \label{dW_i}
 \dot{W}_{i}(\mathbf{x})=-k_0(1-\phi(\text{d}_{\mathcal{O}}(\mathbf{x})))(\mathbf{x}-\mathbf{s}_{i})^{\top}(\mathbf{x}-\mathbf{x}^{\ast}).
 \end{equation}
 Since $(\mathbf{x}-\mathbf{s}_{i})^{\top}(\mathbf{x}-\mathbf{x}^{\ast})\geq0$ for all $\mathbf{x}\in\mathcal{M}_{i}$, one can conclude from \eqref{dW_i} that $\dot{W}_{i}(\mathbf{x})\leq0$ on the set $\mathcal{M}_{i}$ and $\dot{W}_{i}(\mathbf{x})=0$ only occurs at $\mathbf{x}=\mathbf{s}_{i}$. This implies that $\mathcal{M}_{i}$ is a stable manifold of $\mathbf{s}_{i}$ and any initial state on it will converge to $\mathbf{s}_{i}$. Thus, all $\mathbf{x}(0)\in \mathcal{B}(\mathbf{s}_{i},\epsilon^{\prime})\cap\mathcal{M}_{i}\subset\mathcal{M}_{i}$ will converge to $\mathbf{s}_{i}$.

 Case 2): $\mathbf{x}(0)\in\mathcal{N}_{i}\setminus\{\mathbf{s}_i\}$. For all $\mathbf{x}\in\mathcal{N}_{i}\setminus\{\mathbf{s}_i\}$, we have $\text{d}_{\mathcal{O}}(\mathbf{x})=\epsilon$ and $\phi(\text{d}_{\mathcal{O}}(\mathbf{x}))=1$. As such, \eqref{dx_2} becomes
 \begin{equation}
 \label{dx_4}
 \dot{\mathbf{x}}=-k_0\mathbf{Q}(\mathbf{b}(\mathbf{x}))(\mathbf{x}-\mathbf{x}^{\ast}),
 \end{equation}
 where $\mathbf{Q}(\mathbf{b}(\mathbf{x})):=\mathbf{I}_{2}-\mathbf{b}(\mathbf{x})\mathbf{b}^{\top}(\mathbf{x})$ is defined for notational brevity. Consider again the LFC $W_{i}(\mathbf{x})$ in \eqref{W_i}. Then, taking the time derivative of $W_{i}$ along \eqref{dx_4} and noting \eqref{s_i} and the fact that $(\mathbf{x}-\mathbf{c}_{i})^{\top}\mathbf{Q}(\mathbf{b}(\mathbf{x}))=0$, we get
 \begin{align}
 \label{dW_i1}
 \dot{W}_{i}(\mathbf{x})=&~k_0\alpha_i(\mathbf{x}-\mathbf{x}^{\ast})^{\top}\mathbf{Q}(\mathbf{b}(\mathbf{x}))(\mathbf{x}-\mathbf{x}^{\ast}) \nonumber \\
 &-k_0(1-\alpha_i)(\mathbf{x}-\mathbf{c}_{i})^{\top}\mathbf{Q}(\mathbf{b}(\mathbf{x}))(\mathbf{x}-\mathbf{x}^{\ast}) \nonumber\\
 =&~k_0\alpha_i(\mathbf{x}-\mathbf{x}^{\ast})^{\top}\mathbf{Q}(\mathbf{b}(\mathbf{x}))(\mathbf{x}-\mathbf{x}^{\ast}) \nonumber\\
 =&~k_0\alpha_i(1-\cos\psi)\|\mathbf{x}-\mathbf{x}^{\ast}\|^{2}>0,
 \end{align}
 where $\psi$ is the angle between the vectors $\mathbf{b}(\mathbf{x})$ and $\bm{\kappa}_0(\mathbf{x})$ and satisfies $0<\psi<\pi/2$ for all $\mathbf{x}\in\mathcal{N}_{i}\setminus\{\mathbf{s}_i\}$. Further consider the following LFC:
 \begin{equation}
 \label{V_i}
 V_{i}(\mathbf{x})=\frac{1}{2}\|\mathbf{x}-\mathbf{c}_{i}\|^{2},
 \end{equation}
 whose time derivative along \eqref{dx_4} is given by
 \begin{equation}
 \label{dV_i}
 \dot{V}_{i}(\mathbf{x})=-k_0(\mathbf{x}-\mathbf{c}_{i})^{\top}\mathbf{Q}(\mathbf{b}(\mathbf{x}))(\mathbf{x}-\mathbf{x}^{\ast})=0,
 \end{equation}
 showing that for all $\mathbf{x}\in\mathcal{N}_{i}\setminus\{\mathbf{s}_{i}\}$, the control vector is tangent to the boundary of $\mathcal{O}_{i}^{\epsilon}$, thus directing the robot to move along $\partial\mathcal{O}_{i}^{\epsilon}$. In view of \eqref{dW_i1} and \eqref{dV_i}, we can conclude that if $\mathbf{x}(0)\in\mathcal{N}_{i}\setminus{\mathbf{s}_{i}}$, then $\mathbf{x}$ will keep away from $\mathbf{s}_{i}$ along $\partial\mathcal{O}_{i}^{\epsilon}$ until it leaves the set $\mathcal{B}(\mathbf{s}_{i},\epsilon^{\prime})$.

 Case 3): $\mathbf{x}(0)\in\mathcal{F}_{i}\setminus(\mathcal{M}_{i}\cup\mathcal{N}_{i})$. Actually, for any $\mathbf{x}\in\mathcal{F}_{i}\setminus(\mathcal{M}_{i}\cup\mathcal{N}_{i})$, it holds that $\phi(\text{d}_{\mathcal{O}}(\mathbf{x}))\in(0,1)$, and $\bf{\kappa}_{0}(\mathbf{x})$ and $\mathbf{b}(\mathbf{x})$ are not collinear. Moreover, we can decompose $\mathbf{h}(\mathbf{x})=\mathbf{\Pi}(\mathbf{x})\bm{\kappa}_{0}(\mathbf{x})$ into two components: the component $\mathbf{h}(\mathbf{x})_{\parallel}=\mathbf{b}(\mathbf{x})\mathbf{b}^{\top}(\mathbf{x})\mathbf{h}(\mathbf{x})$ parallel to $\mathbf{b}(\mathbf{x})$ and the component $\mathbf{h}(\mathbf{x})_{\perp}=(\mathbf{I}_{2}-\mathbf{b}(\mathbf{x})\mathbf{b}^{\top}(\mathbf{x}))\mathbf{h}(\mathbf{x})$ perpendicular to $\mathbf{b}(\mathbf{x})$, as shown in Fig. \ref{figA}. With \eqref{Pi_x} in mind, one can verify that
 \begin{align}
 \label{eq39}
 \mathbf{b}^{\top}(\mathbf{x})\bm{\tau}(\mathbf{x})_{\parallel}=&~\mathbf{b}^{\top}(\mathbf{x})\bm{\tau}(\mathbf{x})\nonumber \\
 =&~(1-\phi(\text{d}_{\mathcal{O}}(\mathbf{x})))\mathbf{b}^{\top}(\mathbf{x})\bm{\kappa}_{0}(\mathbf{x})>0,
 \end{align}
 which implies that $\mathbf{h}(\mathbf{x})_{\parallel}$ and the bearing vector $\mathbf{b}(\mathbf{x})$ have the same direction, causing $\mathbf{h}(\mathbf{x})_{\parallel}$ to point towards $\mathbf{c}_{i}$. A straightforward calculation yields $(\mathbf{I}_{2}-\mathbf{b}(\mathbf{x})\mathbf{b}^{\top}(\mathbf{x})\mathbf{\Pi}(\mathbf{x})=\mathbf{Q}(\mathbf{b}(\mathbf{x}))$, which allows us to express $\mathbf{h}(\mathbf{x})_{\perp}=\mathbf{Q}(\mathbf{b}(\mathbf{x}))\bm{\kappa}_{0}(\mathbf{x})$. Under $\mathbf{h}(\mathbf{x})_{\perp}$, \eqref{dx_2} turns to be \eqref{dx_4}, which results in $\dot{W}_{i}(\mathbf{x})>0$ as seen in \eqref{dW_i1}. Thus, $\mathbf{h}(\mathbf{x})_{\perp}$ generates a velocity component tangent to $\mathbf{b}$, causing $\mathbf{x}$ to move away from $\mathbf{s}_{i}$. Since $\mathbf{h}(\mathbf{x})$ and the vector $\mathbf{s}_{i}-\mathbf{x}$ are situated on opposite sides of $\mathbf{b}(\mathbf{x})$ (see Fig. \ref{figA}), $\mathbf{x}$ cannot move toward $\mathbf{s}_{i}$ on the set $\mathcal{F}_{i}\setminus(\mathcal{M}_{i}\cup\mathcal{N}_{i})$. From Fig. \ref{figA}, one can intuitively observe that $\mathbf{h}(\mathbf{x})$ guides $\mathbf{x}$ towards the boundary of $\mathcal{O}_{i}^{\epsilon}$ (attributed to $\mathbf{h}(\mathbf{x})_{\parallel})$, while simultaneously keeping away from the manifold $\mathcal{M}_{i}$ (attributed to $\mathbf{h}(\mathbf{x})_{\perp}$). Therefore, for any $\mathbf{x}(0)\in\mathcal{F}_{i}\setminus(\mathcal{M}_{i}\cup\mathcal{N}_{i})$, the solution of \eqref{dx} will either directly leaves the ball $\mathcal{B}(\mathbf{s}_{i},\epsilon^{\prime})$, or first converges to the set $\mathcal{N}_{i}\setminus\{\mathbf{s}_{i}\}$ and then leaves the ball $\mathcal{B}(\mathbf{s}_{i},\epsilon^{\prime})$ along $\partial\mathcal{O}_{i}^{\epsilon}$ (as shown in Case 2). 

 The three cases above demonstrate that each stationary point $\mathbf{s}_{i}$ ($i\in\mathbb{I}$) is an unstable fixed point, but there exists one line of initial conditions, namely the stable manifold $\mathcal{M}_{i}$, that is attracted to $\mathbf{s}_{i}$. Note that $\mathcal{M}_{i}$ is a 1-dimensional manifold with boundary and thus has zero measure \cite{milnor1997topology}. As such, $\mathbf{x}^{\ast}$ is an almost globally asymptotically stable equilibrium, with a basin of attraction consisting of the free space $\mathcal{X}_{\epsilon}$, except for a set of measure zero. Furthermore, as $\mathbf{x}^{\ast}\in\text{int}(\mathcal{X}_{\epsilon})$, there certainly exists $r^\ast>0$ such that $\text{d}_{\mathcal{O}}(\mathbf{x})>\epsilon^{\ast}$ for all $\mathbf{x}\in\mathcal{B}(\mathbf{x}^\ast,r^{\ast})$. On the ball $\mathcal{B}(\mathbf{x}^\ast,r^{\ast})$, the kinematics reduces to
 \begin{equation}
 \label{dx_5}
 \dot{\mathbf{x}}=-k_0(\mathbf{x}-\mathbf{x}^{\ast}),
 \end{equation}
 indicating that the local exponential stability of $\mathbf{x}=\mathbf{x}^\ast$. $\hfill \blacksquare$



%
 
%


\bibliographystyle{IEEEtran}
\bibliography{ref}

\begin{thebibliography}{10}
\providecommand{\url}[1]{#1}
\csname url@samestyle\endcsname
\providecommand{\newblock}{\relax}
\providecommand{\bibinfo}[2]{#2}
\providecommand{\BIBentrySTDinterwordspacing}{\spaceskip=0pt\relax}
\providecommand{\BIBentryALTinterwordstretchfactor}{4}
\providecommand{\BIBentryALTinterwordspacing}{\spaceskip=\fontdimen2\font plus
\BIBentryALTinterwordstretchfactor\fontdimen3\font minus
  \fontdimen4\font\relax}
\providecommand{\BIBforeignlanguage}[2]{{%
\expandafter\ifx\csname l@#1\endcsname\relax
\typeout{** WARNING: IEEEtran.bst: No hyphenation pattern has been}%
\typeout{** loaded for the language `#1'. Using the pattern for}%
\typeout{** the default language instead.}%
\else
\language=\csname l@#1\endcsname
\fi
#2}}
\providecommand{\BIBdecl}{\relax}
\BIBdecl

\bibitem{klancar2017wheeled}
G.~Klancar, A.~Zdesar, S.~Blazic, and I.~Skrjanc, \emph{Wheeled mobile
  robotics: from fundamentals towards autonomous systems}.\hskip 1em plus 0.5em
  minus 0.4em\relax Butterworth-Heinemann, 2017.

\bibitem{chu2022feedback}
X.~Chu, R.~Ng, H.~Wang, and K.~W.~S. Au, ``Feedback control for collision-free
  nonholonomic vehicle navigation on {SE (2)} with null space circumvention,''
  \emph{IEEE/ASME Transactions on Mechatronics}, vol.~27, no.~6, pp.
  5594--5604, 2022.

\bibitem{lumelsky2005sensing}
V.~J. Lumelsky, \emph{Sensing, intelligence, motion: how robots and humans move
  in an unstructured world}.\hskip 1em plus 0.5em minus 0.4em\relax John Wiley
  \& Sons, 2005.

\bibitem{berkane2021navigation}
S.~Berkane, ``Navigation in unknown environments using safety velocity cones,''
  in \emph{2021 American Control Conference (ACC)}.\hskip 1em plus 0.5em minus
  0.4em\relax IEEE, 2021, pp. 2336--2341.

\bibitem{khatib1986real}
O.~Khatib, ``Real-time obstacle avoidance for manipulators and mobile robots,''
  \emph{The International Journal of Robotics Research}, vol.~5, no.~1, pp.
  90--98, 1986.

\bibitem{kim1992real}
J.-O. Kim and P.~Khosla, ``Real-time obstacle avoidance using harmonic
  potential functions,'' \emph{IEEE Transactions on Robotics and Automation},
  vol.~8, no.~3, pp. 338--349, 1992.

\bibitem{valbuena2012hybrid}
L.~Valbuena and H.~G. Tanner, ``Hybrid potential field based control of
  differential drive mobile robots,'' \emph{Journal of Intelligent \& Robotic
  Systems}, vol.~68, pp. 307--322, 2012.

\bibitem{li2021optimization}
H.~Li, W.~Liu, C.~Yang, W.~Wang, T.~Qie, and C.~Xiang, ``An optimization-based
  path planning approach for autonomous vehicles using the dynefwa-artificial
  potential field,'' \emph{IEEE Transactions on Intelligent Vehicles}, vol.~7,
  no.~2, pp. 263--272, 2021.

\bibitem{singletary2021comparative}
A.~Singletary, K.~Klingebiel, J.~Bourne, A.~Browning, P.~Tokumaru, and A.~Ames,
  ``Comparative analysis of control barrier functions and artificial potential
  fields for obstacle avoidance,'' in \emph{2021 IEEE/RSJ International
  Conference on Intelligent Robots and Systems (IROS)}.\hskip 1em plus 0.5em
  minus 0.4em\relax IEEE, 2021, pp. 8129--8136.

\bibitem{srinivasan2020control}
M.~Srinivasan and S.~Coogan, ``Control of mobile robots using barrier functions
  under temporal logic specifications,'' \emph{IEEE Transactions on Robotics},
  vol.~37, no.~2, pp. 363--374, 2021.

\bibitem{chen2020guaranteed}
Y.~Chen, A.~Singletary, and A.~D. Ames, ``Guaranteed obstacle avoidance for
  multi-robot operations with limited actuation: A control barrier function
  approach,'' \emph{IEEE Control Systems Letters}, vol.~5, no.~1, pp. 127--132,
  2020.

\bibitem{rimon1992exact}
E.~Rimon and D.~Koditschek, ``Exact robot navigation using artificial potential
  functions,'' \emph{IEEE Transactions on Robotics and Automation}, vol.~8,
  no.~5, pp. 501--518, 1992.

\bibitem{erke2020improved}
S.~Erke, D.~Bin, N.~Yiming, Z.~Qi, X.~Liang, and Z.~Dawei, ``An improved
  {A}-star based path planning algorithm for autonomous land vehicles,''
  \emph{International Journal of Advanced Robotic Systems}, vol.~17, no.~5, p.
  1729881420962263, 2020.

\bibitem{noreen2016optimal}
I.~Noreen, A.~Khan, and Z.~Habib, ``Optimal path planning using rrt* based
  approaches: a survey and future directions,'' \emph{International Journal of
  Advanced Computer Science and Applications}, vol.~7, no.~11, pp. 97--107,
  2016.

\bibitem{tu2003genetic}
J.~Tu and S.~X. Yang, ``Genetic algorithm based path planning for a mobile
  robot,'' in \emph{2003 IEEE International Conference on Robotics and
  Automation (Cat. No. 03CH37422)}, vol.~1.\hskip 1em plus 0.5em minus
  0.4em\relax IEEE, 2003, pp. 1221--1226.

\bibitem{costa2019survey}
M.~M. Costa and M.~F. Silva, ``A survey on path planning algorithms for mobile
  robots,'' in \emph{2019 IEEE International Conference on Autonomous Robot
  Systems and Competitions (ICARSC)}.\hskip 1em plus 0.5em minus 0.4em\relax
  IEEE, 2019, pp. 1--7.

\bibitem{yan2022hierarchical}
Y.~Yan, L.~Peng, J.~Wang, H.~Zhang, T.~Shen, and G.~Yin, ``A hierarchical
  motion planning system for driving in changing environments: Framework,
  algorithms, and verifications,'' \emph{IEEE/ASME Transactions on
  Mechatronics}, vol.~28, no.~3, pp. 1303--1314, 2023.

\bibitem{lumelsky1990incorporating}
V.~J. Lumelsky and T.~Skewis, ``Incorporating range sensing in the robot
  navigation function,'' \emph{IEEE Transactions on Systems, Man, and
  Cybernetics}, vol.~20, no.~5, pp. 1058--1069, 1990.

\bibitem{mcguire2019comparative}
K.~N. McGuire, G.~C. de~Croon, and K.~Tuyls, ``A comparative study of bug
  algorithms for robot navigation,'' \emph{Robotics and Autonomous Systems},
  vol. 121, p. 103261, 2019.

\bibitem{arslan2019sensor}
O.~Arslan and D.~E. Koditschek, ``Sensor-based reactive navigation in unknown
  convex sphere worlds,'' \emph{The International Journal of Robotics
  Research}, vol.~38, no. 2-3, pp. 196--223, 2019.

\bibitem{huber2019avoidance}
L.~Huber, A.~Billard, and J.-J. Slotine, ``Avoidance of convex and concave
  obstacles with convergence ensured through contraction,'' \emph{IEEE Robotics
  and Automation Letters}, vol.~4, no.~2, pp. 1462--1469, 2019.

\bibitem{zhai2019adaptive}
J.-Y. Zhai and Z.-B. Song, ``Adaptive sliding mode trajectory tracking control
  for wheeled mobile robots,'' \emph{International Journal of Control},
  vol.~92, no.~10, pp. 2255--2262, 2019.

\bibitem{zheng2024adaptive}
Y.~Zheng, J.~Zheng, K.~Shao, H.~Zhao, H.~Xie, and H.~Wang, ``Adaptive
  trajectory tracking control for nonholonomic wheeled mobile robots: A barrier
  function sliding mode approach,'' \emph{IEEE/CAA Journal of Automatica
  Sinica}, vol.~11, no.~4, pp. 1007--1021, 2024.

\bibitem{kim2023energy}
Y.~Kim and T.~Singh, ``Energy-time optimal trajectory tracking control of
  wheeled mobile robots,'' \emph{IEEE/ASME Transactions on Mechatronics},
  vol.~29, no.~2, pp. 1283--1294, 2024.

\bibitem{zhou2023homogeneity}
Y.~Zhou, H.~R{\'\i}os, M.~Mera, A.~Polyakov, G.~Zheng, and A.~Dzul,
  ``Homogeneity-based control strategy for trajectory tracking in perturbed
  unicycle mobile robots,'' \emph{IEEE Transactions on Control Systems
  Technology}, vol.~32, no.~1, pp. 274--281, 2024.

\bibitem{siciliano2009mobile}
B.~Siciliano, L.~Sciavicco, L.~Villani, and G.~Oriolo, \emph{Robotics:
  Modelling, Planning and Control}.\hskip 1em plus 0.5em minus 0.4em\relax
  Springer, 2009.

\bibitem{tran2020finite}
D.~Tran and T.~Yucelen, ``Finite-time control of perturbed dynamical systems
  based on a generalized time transformation approach,'' \emph{Systems \&
  Control Letters}, vol. 136, p. 104605, 2020.

\bibitem{bouligand1932introduction}
G.~Bouligand, ``Introduction {\`a} la g{\'e}om{\'e}trie infinit{\'e}simale
  directe,'' \emph{(No Title)}, 1932.

\bibitem{nagumo1942lage}
M.~Nagumo, ``{\"U}ber die lage der integralkurven gew{\"o}hnlicher
  differentialgleichungen,'' \emph{Proceedings of the Physico-Mathematical
  Society of Japan. 3rd Series}, vol.~24, pp. 551--559, 1942.

\bibitem{lee2022manifolds}
J.~M. Lee, \emph{Manifolds and Differential Geometry}.\hskip 1em plus 0.5em
  minus 0.4em\relax American Mathematical Society, 2022, vol. 107.

\bibitem{shao2023fault}
X.~Shao, Q.~Hu, Y.~Shi, and Y.~Zhang, ``Fault-tolerant control for full-state
  error constrained attitude tracking of uncertain spacecraft,''
  \emph{Automatica}, vol. 151, p. 110907, 2023.

\bibitem{wang2024hybrid}
M.~Wang and A.~Tayebi, ``Hybrid feedback for affine nonlinear systems with
  application to global obstacle avoidance,'' \emph{IEEE Transactions on
  Automatic Control}, vol.~69, no.~8, pp. 5546--5553, 2024.

\bibitem{arvin2019mona}
F.~Arvin, J.~Espinosa, B.~Bird, A.~West, S.~Watson, and B.~Lennox, ``Mona: an
  affordable open-source mobile robot for education and research,''
  \emph{Journal of Intelligent \& Robotic Systems}, vol.~94, pp. 761--775,
  2019.

\bibitem{chaney1990piecewise}
R.~W. Chaney, ``Piecewise $c^k$ functions in nonsmooth analysis,''
  \emph{Nonlinear Analysis: Theory, Methods \& Applications}, vol.~15, no.~7,
  pp. 649--660, 1990.

\bibitem{2002Nonlinear}
H.~K. Khalil, \emph{Nonlinear Systems (3rd edn)}.\hskip 1em plus 0.5em minus
  0.4em\relax NJ: Prentice Hall, 2002.

\bibitem{milnor1997topology}
J.~W. Milnor and D.~W. Weaver, \emph{Topology from the differentiable
  viewpoint}.\hskip 1em plus 0.5em minus 0.4em\relax Princeton University
  Press, 1997.

\end{thebibliography}

\newpage

 
\vspace{11pt}


%

\vfill

\end{document}